\pdfoutput=1
\PassOptionsToPackage{usenames,dvipsnames}{xcolor}
\documentclass[11pt]{article}

\newif\ifarxiv
\arxivtrue

\ifarxiv
    \usepackage{acl}
\else
    \usepackage[review]{acl}
\fi

\usepackage{times}
\usepackage{latexsym}
\usepackage{amsfonts}
\usepackage[T1]{fontenc}

\usepackage[utf8]{inputenc}

\usepackage{inconsolata} 

\usepackage{microtype}
\usepackage{sec/package}
\usepackage[dvipsnames]{xcolor} 
\usepackage{tabularx} 
\newcommand{\data}[1]{\textsc{AI Scholar}\xspace}

\title{

Voices of \textit{Her}:
Analyzing Gender Differences in the AI Publication World

}

\author{
Yiwen Ding\textsuperscript{\rm 1,}\thanks{Equal contribution.
} \quad
Jiarui Liu\textsuperscript{\rm 6,}\samethanks \quad
Zhiheng Lyu\textsuperscript{\rm 2,}\samethanks \quad
Kun Zhang\textsuperscript{\rm 6,7} \quad
Bernhard Schölkopf\textsuperscript{\rm 3}
\\  
{\bf
Zhijing Jin\textsuperscript{\rm 3,4,5}\thanks{Equal supervision.} \quad
Rada Mihalcea\textsuperscript{\rm 1,}\samethanks
}
\\
\textsuperscript{\rm 1}University of Michigan
{}
\textsuperscript{\rm 2}University of Hong Kong
\\
\textsuperscript{\rm 3}Max Planck Institute for Intelligent Systems, Tuebingen, Germany
\\
\textsuperscript{\rm 4}University of Toronto
{}
\textsuperscript{\rm 5}Vector Institute
{}
\textsuperscript{\rm 6}CMU
{}
\textsuperscript{\rm 7}MBZUAI
\vspace{.3em}
\\
\texttt{
yvonneding99@gmail.com
{}
jiarui@cmu.edu 
{}
zhiheng.lyu.cs@gmail.com
}
\\
\texttt{
mihalcea@umich.edu
{}
zjin@cs.toronto.edu
}
}

\begin{document}
\maketitle
\begin{abstract}

While several previous studies have analyzed gender bias in  research, we are still missing a comprehensive analysis of gender differences in the AI community, covering diverse topics and different development trends. Using the \data{} dataset of 78K researchers in the field of AI, we identify several gender differences: (1) Although female researchers tend to have fewer overall citations than males, this citation difference does not hold for all academic-age groups; (2) There exist large gender homophily in co-authorship on AI papers; (3) Female first-authored papers show distinct linguistic styles, such as longer text, more positive emotion words, and more catchy titles than male first-authored papers. Our analysis provides a window into the current demographic trends in our AI community, and encourages more gender equality and diversity in the future.\footnote{Our code and data
\arxivfalse
\ifarxiv
are at \url{https://github.com/causalNLP/ai-scholar-gender}.
\else
have been uploaded to the submission system, and will be open-sourced upon acceptance.
\fi
}
\end{abstract}

\section{Introduction}
\ifarxiv
Motivated by the spirit of the ACL Year-Round Mentorship Program\footnote{\url{https://mentorship.aclweb.org/}} to support junior researchers to understand how a career path in NLP is, we want to answer this question technically, namely, what are the causal factors for academic success.
\fi


Although nearly half of the world population is female \cite{owidgenderratio}, the proportion of female researchers in science fields is often disproportionately smaller \cite{robnett2016gender, hand2017exploring}.
Specifically, in the research community of AI, we find that female researchers constitute only \textbf{17.99\%} of all the scholars in the field of AI with more than 100 citations, as collected
in the \data{} dataset
\cite{jin2022ai}. This fraction is even smaller in some subdomains of AI such as computer vision (CV), where only 15.64\% of the researchers are females.

Despite this strong gender gap among AI researchers, comprehensive research on the state of the field is yet to be conducted. Most existing research work has been done on certain subdomains of AI, such as the NLP community \cite{vogel-jurafsky-2012-said,schluter-2018-glass,mohammad-2020-gender}, or has addressed research aspects such as the values listed in  top-cited papers \cite{birhane2021values}. To the best of our knowledge, this study is the first to conduct an up-to-date AI community-wise comprehensive analysis since the study of \citet{stathoulopoulos2019gender}.

\begin{figure}[t]
    \centering
    \includegraphics[width=\columnwidth]{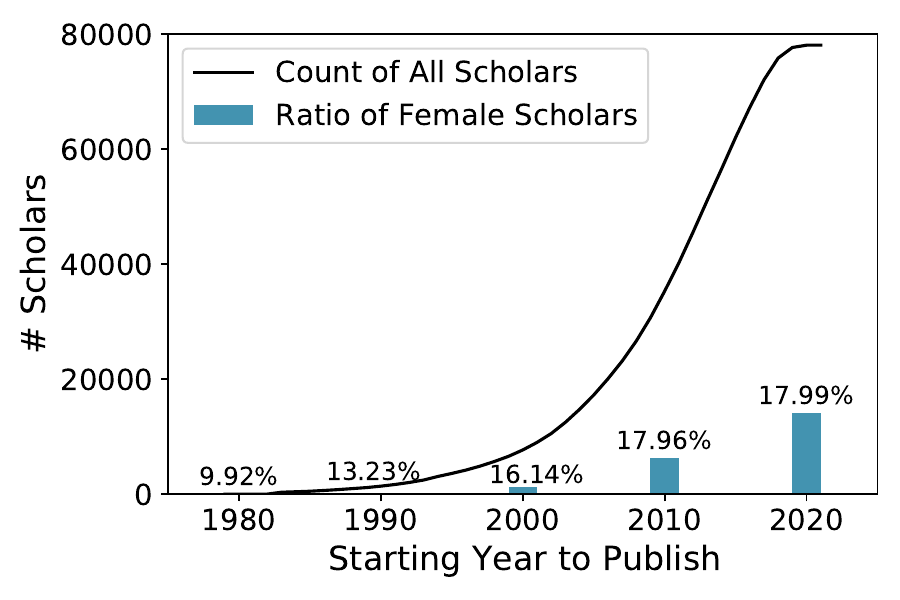}
    \caption{Cumulative number of AI scholars and female scholars ratio vs. years to publish the first paper. Female scholars are taking an increasing percentage of all AI scholars, but the ratio is still small (around 18\%).}
    \label{fig:scholar_pub_trend}
\end{figure}
In this paper, we look into distinct features of the female subgroup in the AI community, and conduct comprehensive statistical analyses from a diverse range of perspectives: basic scholar profile statistics, citation trends, coauthorship, and linguistic styles of papers.
The main findings from our study are as follows:
\begin{enumerate}
    \item Although female AI researchers tend to have fewer overall citations than males, this citation difference does not always hold for all academic-age groups or all time stages in one's career.
    \item There exists large homophily in genders of the first author, last author, and the majority of the authors on AI papers, such as a high correlation between male last authors and the majority of authors being male. This gender homophily pattern in AI echoes the observation of \citet{schluter-2018-glass} in NLP.
    \item Female first-authored papers tend to have distinct linguistic patterns such as more words about positive emotion, longer text, and more catchy titles.
\end{enumerate}


Our findings contribute suggestions and supporting evidence to future AI community organizers or individuals who want to push for informed community changes.

\section{Data Collection and Cleaning}

\paragraph{\data{}.}
We use all the scholar information from the most recent collection of researchers in the field of AI, i.e., the \data{} dataset \cite{jin2022ai},\footnote{Dataset is available at \url{https://github.com/causalNLP/AI-Scholar}} which contains all the scholars in the field of AI with at least 100 citations according to Google Scholar. The data consists of 78K scholars with tags related to  AI such as artificial intelligence (AI) and machine learning (ML), or subdomains of AI such as computer vision (CV) and natural language processing (NLP). It only includes scholars with at least 100 citations, an approximate cut-off for the long-tail since it is not feasible to include all scholar profiles. We discuss the limitations of using this dataset in \cref{sec:limitations}. Throughout the paper, we use the term ``\textit{AI researchers}'' to denote the set of scholars in the \data{} dataset.

For each AI researcher, the \data{} dataset collects information such as the name, affiliation, up to five domain tags, total citations, citations by year, and all their papers with title, year, and the number of citations.

Since the total number of papers is massive (2.8M papers for the 78K AI researchers), we use the random subset of papers provided by \citet{jin2022ai}.
They collect 100K papers with detailed information such as abstracts and full names of all the coauthors. Among the 100K papers with detailed information, we further filter out papers with empty abstracts and keep 91K
papers, which we denote as ``\textit{AI papers}'' in our analysis.

\paragraph{Identifying Female Researchers.}
Since the focus of this paper is to analyze the female subgroup in the AI community, we have to find a way to identify AI researchers that are female. Admittedly, this is a daunting task  due to two main concerns. First, gender is a continuum that goes beyond the male/female binary distinction. Second, there are no computational methods to identify the gender of a researcher that are perfectly correct and perfectly ethical. A possible way is to collect as many self-reports of gender as possible, but this method will be largely time-consuming on the scale of 78K, and also might lead to a large selection bias in the data, since the collected responses might be of a small number and not an \textit{i.i.d.} subset of the entire data. After balancing all the ethical and practical concerns, we decided to follow the practice from  \citet{mohammad-2020-gender}, who classified gender by collecting first names that correspond to male and female genders more than 95\% of the time in the merged records of the US Social Security Administration's published database of names and genders along with the PubMed authors with genders, as well as using the hand-labeled author genders by \citet{vogel-jurafsky-2012-said} to correct for  wrongly classified names. Using this conservative but ethical approach, we obtain 7,036 female authors and 32,074 male authors from the 78,066 AI researchers, and leave the author names that cannot be classified as ``unclassified.''

We acknowledge that the name-gender records that we use have limited representations of names from all cultural backgrounds.
In our paper, we make an effort to keep the errors modular, such that future work can use our analysis framework on a more accurate set of female AI researchers to produce more accurate insights.

\paragraph{Population with Unclassified Gender.}
Since we choose to stick with our ethical standards, including not using any name- or photo-based classifier, we have to leave out a large set of AI researchers whose gender cannot be identified. 
To address this, we would like to frame the scope our analysis to this subset of AI researchers whose gender can be identified. And also we believe that despite this limitation, this study is still more meaningful to the community than not drawing any conclusions. In \cref{appd:unclassified}, we analyze the coverage and properties of this subset.

\section{Analysis of Scholar Profiles}

We first analyze the basic scholar profiles to compare the general statistics with those of the female subgroup. Our analyses answer the following questions: (1) What percentage of female scholars are there in AI and in each subdomain? (2) What are the scholarly indices of the female researchers doing in AI? And (3) How do the analyses differ if we consider the scholarly trends before and after 2012 -- a year that corresponds approximately to the time when deep learning started to become widely used (among others, it is the publication year of AlexNet \cite{krizhevsky2012imagenet}) -- w.r.t. different academic age groups, and in academia vs. industry?
Note that for the scope of this paper, we focus on overall trends,
and we encourage future work to dive into causal analysis.


\subsection{Female Percentage}

We first check the size of the female subgroup in AI and various subdomains of AI.

\begin{figure}[ht]
    \centering
    \includegraphics[width=\columnwidth]{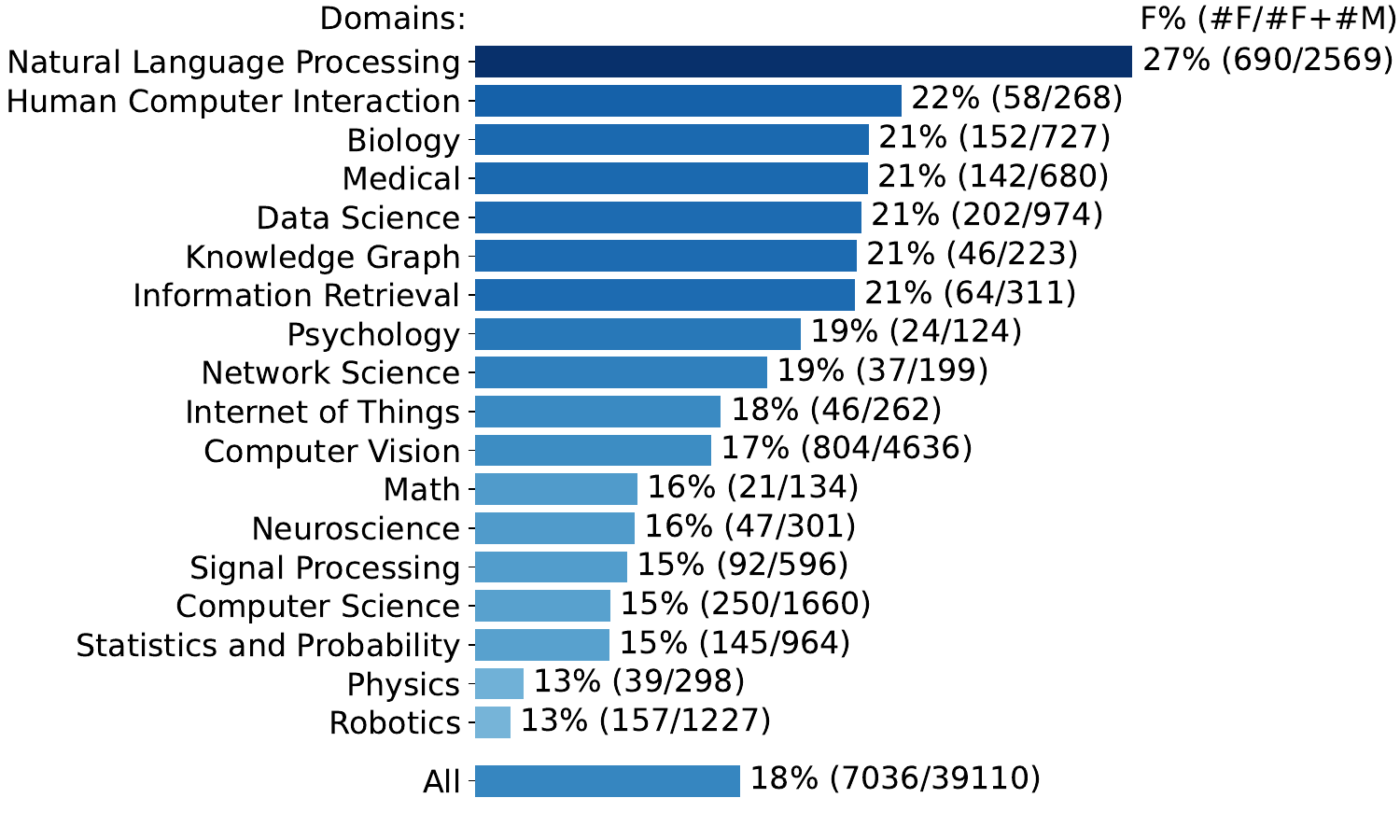}
    \caption{Female scholar percentage (F\%) by subdomains of AI collected from Google Scholar profiles.
    }
    \label{fig:domain}
\end{figure}
In \cref{fig:domain}, we can see that there are 17.99\% female scholars among all AI researchers with classified gender, and this percentage varies across the different subdomains of AI that scholars self-label on their Google Scholar profiles.
The representation of females is relatively more pronounced in areas such as natural language processing (27\%) and human-computer interaction (22\%), and less seen in areas such as physics and robotics, both with only 13\%.
We discuss the experimental details in \cref{appd:domain_tag_cleaning}, including how we manually clean and cluster these tags as well as count normalization.
Note that to get an informative percentage of female researchers (denoted as ``F\%''), when we calculate the percentages throughout this paper, we consider female scholars among all scholars whose gender is classified, because the non-trivial size of the unclassified group may make the percentage of female scholars look disproportionally small, thus not very informative for understanding the statistics.

\subsection{Profile Statistics}\label{sec:profile_stats}
Next, we look into the profiles of AI researchers and calculate overall scholarly statistics, reported in \cref{tab:profile_num_features}.


\begin{table}[ht]
    \centering
    \small
    \setlength\tabcolsep{2pt}
    \resizebox{\columnwidth}{!}{     \begin{tabular}{p{1.5cm}lccccccc}
    \toprule
    & & Avg & Min & 25th & 50th & 75th & Max \\
    \midrule
    Citations: & All & 2,129.54  &
    100 
    & 214 & 475 & 1,345 & 533,757\\
    & F. & 1,762.11  &
    100 
    & 197 & 414 & 1,165 & 209,549 \\ \hline
    h-Index: & All & 14.03 & 1 & 7 & 10 & 16 & 266\\
    & F. & 13.25 & 1 & 6 & 9 & 15 & 211 \\ \hline
    \# Papers: & All & 67.44 & 1 & 17 & 32 & 68 & 3,000\\
    & F. & 60.20 & 1 & 16 & 29 & 64 & 2,125\\ \hline
    AcadAge: & All & 16.89 & 2 & 10 & 14 & 20 & 73\\
    & F. & 16.33 & 2 & 10 & 14 & 20 & 73\\ \hline
    Active Yrs.: & All & 15.47 & 1 & 8 & 13 & 19 & 73\\
    & F. & 14.87 & 1 & 8 & 12 & 19 & 72\\ \hline
     F. Coauthor & All & 9.58 &0 &5.21 & 8.70 & 12.50 & 48.72  \\
    (\%): & F. & 14.53 & 1.64 & 9.09 & 13.04 & 18.58 & 48.72 \\ \hline

    Academia & All & 60.25 & 0 & 0 & 100 & 100 & 100 \\
    (\%): & F. & 61.79 & 0 & 0 & 100 & 100 & 100 \\ \hline
    Big 10 (\%): & All & 6.61 & 0 & 0 & 0 & 0 & 100 \\
    & F. & 15.97 & 0 & 0 & 0 & 0 & 100 \\
    \bottomrule
    \end{tabular}
   }
    \caption{Statistics of Google Scholar profiles. We compare the statistics for the total population of all
    researchers (``All'') with those of the female subgroup (``F.''), w.r.t. the average; minimum; 25th, 50th, and 75th percentile; and maximum values.
    The reported statistics include citations, h-indices, number of papers (\# Papers), academic age (calculated by subtracting the year of the first paper from the current year), the number of active years (calculated by subtracting the year of the first paper from the year of the last paper), percentage of females among their coauthors (F. Coauthor), proportion of researchers who are affiliated with academia, and the proportion of researchers who are affiliated with the most frequently appearing ten organizations among AI researchers (Big 10).
    See \cref{appd:general_profile} for a more comprehensive table including the standard deviation.
    }
    \label{tab:profile_num_features}
\end{table}


As we can see in \cref{tab:profile_num_features}, the average citation number for female researchers is 1.7K, which is 367 less than that of all AI researchers.
If we look closely at the detailed information on the citation distribution, we can see that this gap may be attributed to the difference in highly cited scholars.
Here, the citation difference is moderate until the 50th percentile, with only --61 difference, but the gap drastically increases in higher percentiles, such as --180 in the 75th percentile, and finally --324K in the maximum.
Across other statistics, we can see a similar trend in the h-index and the number of papers.

The most significant differences can be seen on the percentage of female coauthors.
In \cref{tab:profile_num_features},  average scholars have 9.58\% female coauthors among all coauthors, but female scholars have 14.53\% female coauthors. It is also noteworthy that the percentage of female coauthors for average scholars is less than 10\% until the 50th percentile, which can demonstrate that some fields or coauthorship sub-networks have a very low representation of females.

As some additional notes, we can see that,
on average, females' academic age is slightly younger by 0.56 years.
And,
if we
account for scholars who have stopped publishing by reporting
the number of active years (the number of years between the year that a scholar first published a paper and the year that they published their last paper), then we can see a slightly larger gap of 0.6 years.
This shows a slight trend that females stop publishing a bit earlier.

In the last row of \cref{tab:profile_num_features}, we can see that female researchers are more concentrated in the most frequently-appearing ten organizations: Google, Stanford, CMU, MIT, Amazon, UCB, Microsoft, Facebook, IBM and Apple.
See the implementation details to extract these organizations in \cref{appd:impl_academia_and_orgs}.
There could be many potential explanations, such as that the big organizations have a stronger diversity requirement to bridge the gender gap, or it could be that females who persist in the research field are very talented, among many other possibilities.

\subsection{Varying Views}
After analyzing the overall statistics, we also perform analyses on different subsets of the data: (1) scholarly statistics before and after 2012 when the wide use of deep learning has started; (2) citations by different academic age groups and at different career stages; and (3) academic vs. industry affiliations.

\paragraph{The 2012 AI Wave.}
Since a domain such as AI can go through many ups and downs, we also want to check how the statistics differ before and after the time when deep learning achieved the first round of large empirical success. As a rough estimate, we take the year 2012, when the highly influential paper, AlexNet \cite{krizhevsky2012imagenet} on ImageNet \cite{deng2009imagenet}, was published.

In \cref{tab:newbie_citation}, we compare statistics of AI researchers who published their first paper after 2012 (the post-2012 generation) and before 2012 (the pre-2012 generation). We can see that it takes clearly fewer years for a researcher to reach 100 citations in the post-2012 generation, due to the surge of research on AI after 2012.
Some differences between female scholars among all scholars are slightly larger in the post-2012 generation than in the pre-2012 generation.

\begin{table}[t]
    \centering
    \small
    \setlength\tabcolsep{2pt}
    \resizebox{\columnwidth}{!}{
    \begin{tabular}{llccc}
    \toprule
    & & \multicolumn{1}{c}{
    Post-2012
    } & \multicolumn{1}{c}{
    Pre-2012
    }
    \\
    \midrule
    \# Papers/Yr.: & All & 2.86{\tiny $\pm$2.78} &  3.96{\tiny $\pm$4.32} & \\
    & F. & 2.62{\tiny $\pm$3.12} & 3.77{\tiny $\pm$4.05} \\
    Citations/Paper: & All & 47.93{\tiny $\pm$146.58} & 42.06{\tiny $\pm$89.33} \\
    & F. & 45.02{\tiny $\pm$129.13} & 39.73{\tiny $\pm$91.68} \\
    Yrs.~to Reach 100 Cit. ($\downarrow$): & All & 4.82{\tiny $\pm$1.89}  & 6.77{\tiny $\pm$3.69}\\
    & F. & 5.03{\tiny $\pm$1.94} & 6.93{\tiny $\pm$3.67}\\ \hline
    \textit{Most Cited Paper} \\
    \quad Avg. Citations: & All & 369.41{\tiny $\pm$2208.56} & 692.35{\tiny $\pm$2835.00} \\
    & F. & 316.49{\tiny $\pm$1661.53} & 553.63{\tiny $\pm$1829.00} \\
    \quad Most Common Yr.: & All & 2018 & 2011 \\
    & F. & 2018 & 2011 \\
    \bottomrule
    \end{tabular}
    }
    \caption{Statistics of AI researchers who published their first paper after 2012 (the Post-2012 Generation) and before 2012 (the Pre-2012 Generation). 
    For the most cited paper of each scholar, we list the average citations (Avg. Citations), and the most common year for the most cited paper (Most Common Yr.).}
    \label{tab:newbie_citation}
\end{table}

\paragraph{Academic Age.}
We also explore  the citation differences  across different academic age groups, inspired by the analysis of NLP scholars \citet{mohammad-2020-gender}.
We separate the citations of all scholars and female scholars across two dimensions: each age group (e.g., 0 -- 5, 6 -- 10, \dots), and every 5-year window for each group (e.g., citations by 5th year, 10th year, \dots).

\begin{table}[ht]
    \centering
    \setlength\tabcolsep{0.2pt}
    \resizebox{\columnwidth}{!}{\begin{tabular}{lcccccccc}
    \toprule
    & Cit.~by 5th Yr & By 10th Yr & By 15th Yr & By 20th Yr & By 25th Yr & By 2022 & Total Ratio \\
    AcadAge & All/F & All/F & All/F & All/F & All/F & All/F & All/F \\
    \midrule
    0--5 & \bluebox{200}{200/188} &  &  &  &  &  & 1.06 \\
    6--10 & \bluebox{127}{127/114} & \bluebox{275}{275/248} &  &  &  &  & 1.11 \\
    11-15 & \bluebox{79}{79/72} & \bluebox{300}{300/253} & \bluebox{418}{418/349} &  &  &  & 1.20 \\
    16--20 & \bluebox{79}{81/77} & \bluebox{324}{324/299} & \bluebox{599}{599/544} & \bluebox{724}{724/698} &  &  & 1.04 \\
    21--25 & \bluebox{85}{97/95} & \bluebox{371}{371/366} & \bluebox{793}{793/789} & \bluebox{1029}{1,029/988} & \bluebox{1209}{1,209/1,215} &  & 1.00 \\
    >25 & \bluebox{109}{109/104} & \bluebox{391}{391/383} & \bluebox{891}{891/886} & \bluebox{1519}{1,519/1,473} & \bluebox{1983}{1,983/1,872} & \bluebox{2090}{2,090/1,961} & 1.07 \\
    \bottomrule
    \end{tabular}}
    \caption{Median citations from different academic ages from scholars of different academic age groups (average scholars statistics / female scholars statistics).}
    \label{tab:citation_in_each_academic_age_group}
\end{table}

With this more time-specific view, in \cref{tab:citation_in_each_academic_age_group}, we can see almost equality in several academic age groups with certain time spans, such as the 21 -- 25 academic age group, where the female overall citation by the 25th year (1,215) is even higher than the average (1,209). The citation difference that we see in previous sections can be attributed to more specific age groups and time, such as the 11 -- 15 age group, and the 16 -- 20 age group.

\paragraph{Dropout and Industry.}
Some possible alternative reasons why a scholar has fewer citations could just be a matter of career choice. We want to account for the affiliation difference of scholars (i.e., whether a scholar is in the industry or not) and its correlation with some results that can affect citations (e.g., whether the scholar stops publishing). Therefore, we calculate the correlation between whether a scholar is only affiliated with industry and whether they stopped publishing recently.
Note that we take the year 2018 as an empirical threshold for the recent stop in publication, because some domains may take  longer to publish and 2018 is a relatively reasonable year that avoids the effects of the COVID-19 outbreak.

\begin{table}[ht]
    \centering
    \small
    \begin{tabular}{lcccc}
    \toprule
    & & \multicolumn{2}{c}{Affiliation Is Industry Only}
    \\ \cline{3-4}
    & & No & Yes \\
    \midrule
    \multirow{2}{*}{Stops Publishing:} & No & 41,217 & 25,484 & \\
    & Yes & 2,749 & 3,521 \\
    \bottomrule
    \end{tabular}
    \caption{Contingency table of being in the industry only and stopping publishing since 2018. We get a p-value of $1e$$-169$ by $\chi^2$ test, confirming a strong correlation between being in industry only and stopping publishing.}
    \label{tab:correlation_dropout_industry}
\end{table}

In \cref{tab:correlation_dropout_industry}, we can see that a $\chi^2$ test confirms a strong correlation between a stop in publication and being exclusively affiliated with the industry. When conditioning on all people that keep publishing, the number in academia is almost twice that in the industry. Additionally,  when conditioning on all people that stop publishing, there are 28\% more people in the industry than that in academia.
We include a fine-grained analysis by academic age in \cref{appd:dropout}.

\section{Analysis of Citation Time Series}
To take the analysis one step further, we perform a more fine-grained analysis of the scholar statistics. 


\begin{figure}[t]
\minipage{0.5\columnwidth}
    {\begin{center}
    \small
        {Exponential \leavevmode\hbox to3pt{%
   \pdfliteral{q 0.1 J .35 0 0.2 .35 -2 -2 cm
      10 0 m 25 0 20 20 10 30 c S 10 30 m 11 23 l 19 23 l h B Q}\hss}\quad Cluster}
    \end{center}
    \vspace{-2mm}
    }
  \includegraphics[width=\linewidth]{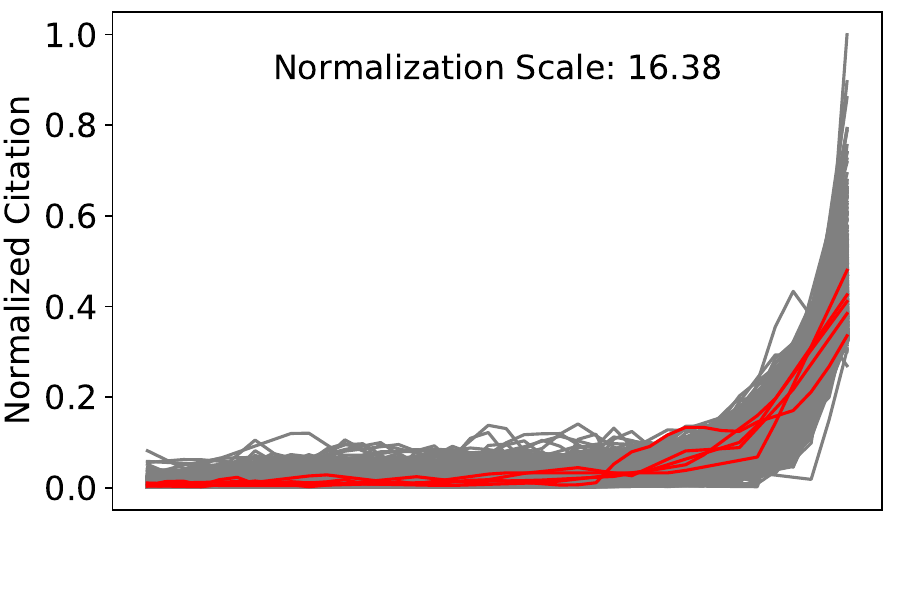}
\endminipage
\hfill
\minipage{0.5\columnwidth}
    {\begin{center}
    \small
        {Linear
        $\nearrow$
        Cluster}
    \end{center}
    \vspace{-2mm}
    }
  \includegraphics[width=\linewidth]{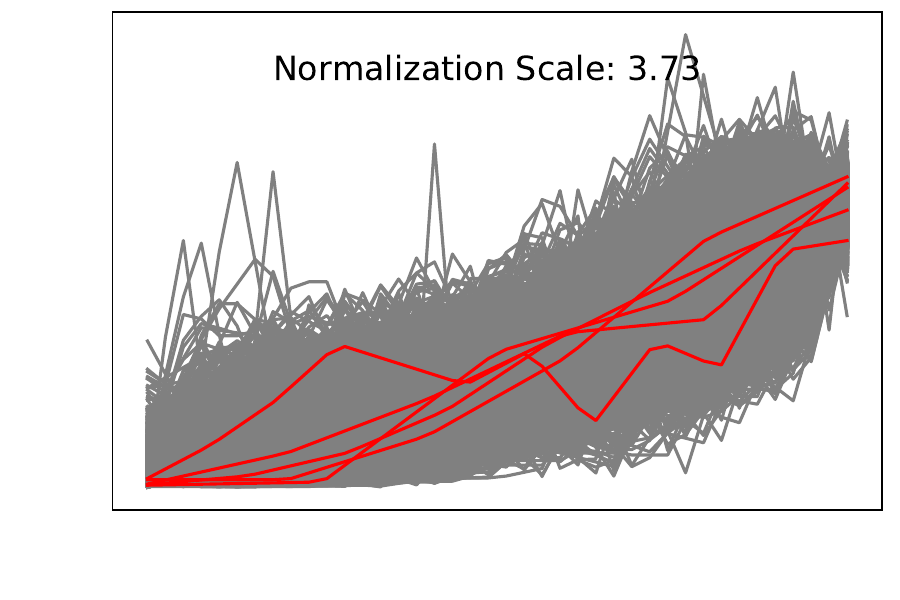}
\endminipage\hfill
\minipage{0.5\columnwidth}%
    {\begin{center}
    \small
        {Stumbling
        $\curvearrowright$ Cluster}
    \end{center}
    \vspace{-2mm}
    }
  \includegraphics[width=\linewidth]{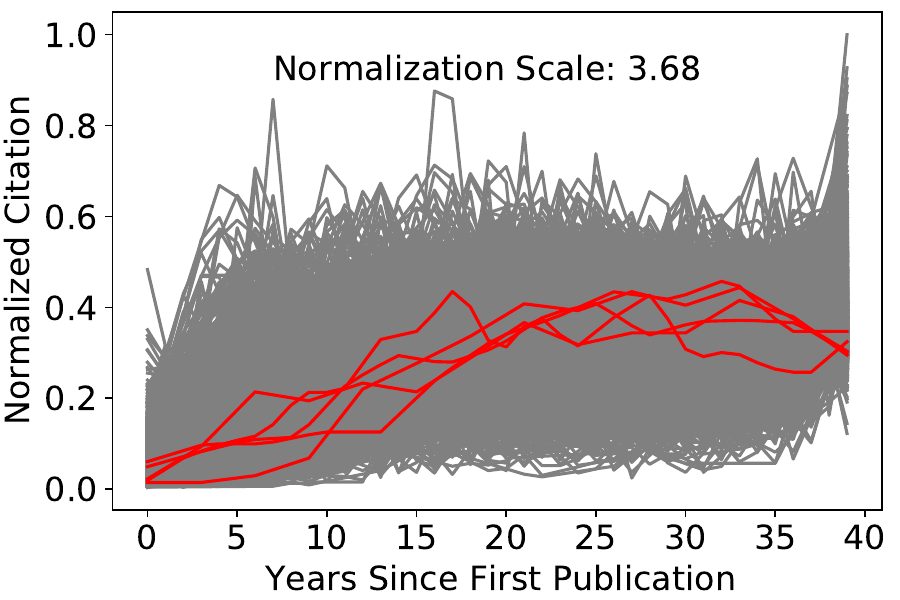}
\endminipage\hfill
\minipage{0.5\columnwidth}
    {\begin{center}
    \small
        {Struggling
        $\rightsquigarrow$
        Cluster}
    \end{center}
    \vspace{-2mm}
    }
  \includegraphics[width=\linewidth]{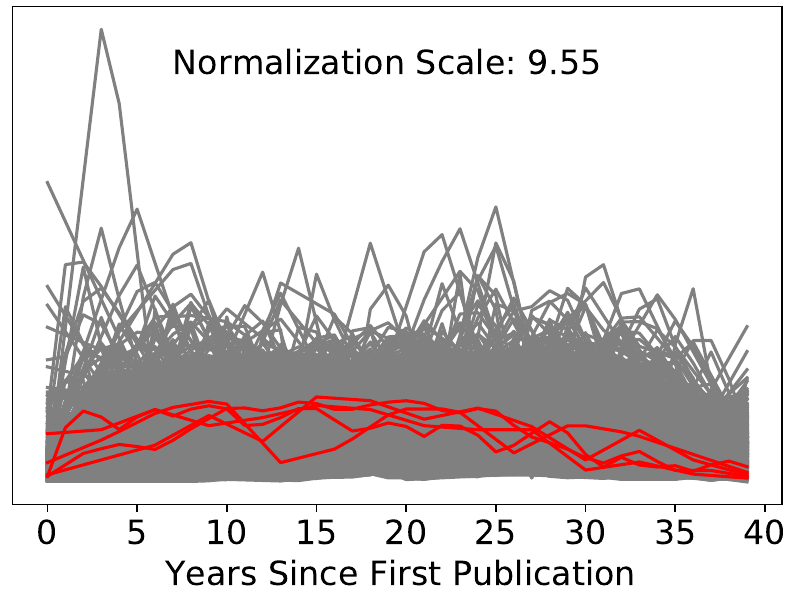}
\endminipage
\caption{
Four main types of time series clusters of AI researchers' citations.}
\label{fig:clusters_citation_by_year}
\end{figure}

\paragraph{Time Series Clustering.}
We are interested in patterns in the scholar citation time series. Inspired by the time series construction by \citet{Tanveer-2018}, we take the  citations-by-year data of all the 78K scholars, normalize them by the average citation number, and linearly interpolate the citation time span to the largest number of active years, so that we can focus on the shape of the citation times series and stay agnostic with respect to different academic ages. We apply K-Means clustering on time series  \cite{JMLR:v21:20-091} and introduce our implementation details in \cref{appd:cluster_time_trends}.


We further manually group the multiple clusters generated by the algorithm into four main types according to human-interpretable shape patterns.
For notation convenience, we manually assign some easy-to-remember names to the four cluster types:
the exponential cluster (\leavevmode\hbox to3pt{%
   \pdfliteral{q 0.1 J .35 0 0.2 .35 -2 -2 cm
      10 0 m 25 0 20 20 10 30 c S 10 30 m 11 23 l 19 23 l h B Q}\hss}\quad),
linear cluster ($\nearrow$), stumbling cluster ($\curvearrowright$), and struggling cluster ($\rightsquigarrow$).
For each type, we visualize a representative cluster in \cref{fig:clusters_citation_by_year}, and plot all machine-identified clusters in \cref{appd:machine-identified}.

\paragraph{Cluster Statistics.}
\begin{table}[t]
    \centering
    \setlength\tabcolsep{2pt}
    \resizebox{\columnwidth}{!}{     \begin{tabular}{llcccccc}
    \toprule
    & & Exponential & Linear & Stumbling & Struggling \\ \midrule
    \# Scholars: & All & 8,008 & 41,698 & 16,857 & 1,565 \\
    & F & 1,535 & 7,831 & 2,818 & 239 \\
    & Ratio (All:F) & 5.22:1 & 5.32:1 & 5.98:1 & 6.56:1 \\
    \midrule
    Citation: & All & 2,472.55 & 2,292.87 & 2,114.19 & 787.80\\
    & F. & 1,569.49 & 1,897.35 & 1,895.13 & 438.15\\
    & Ratio & 1.58:1 & 1.21:1 & 1.12:1 & 1.80:1 \\ \midrule
    h-Index: & All & 16.21 & 14.34 & 13.34 & 9.86\\
    & F. & 14.68 & 13.69 & 12.56 & 8.50\\
    & Ratio & 1.10:1 & 1.05:1 & 1.06:1 & 1.16:1 \\ \midrule
    AcadAge: & All & 17.40 & 14.64 & 19.24 & 21.95\\
    & F. & 16.89 & 14.53 & 18.25 & 19.92\\
    & Ratio & 1.03:1 & 1.01:1 & 1.05:1 & 1.10:1\\
    \midrule
    Stop Pub.: & All & 0.25\% & 2.58\%& 22.98\%& {44.53\%}\\
    & F. & 0.14\% & 2.35\%& 24.45\%& {44.27\%}\\
    & Ratio & 1.79:1 & 1.10:1 & 0.94:1 & 1.01:1 \\
    \bottomrule
    \end{tabular}
    }
    \caption{Scholar statistics in each cluster.
    }
    \label{tab:clusters_gender_gap}
\end{table}

In \cref{tab:clusters_gender_gap}, we can see that the majority of the scholars are in the linear cluster, which is the most common time-series pattern. The exponential pattern is substantially rarer than the linear pattern, only 1/5 by the number of scholars, but with the largest h-indices across all clusters. In the exponential cluster, although female scholars have a higher representation than in other clusters, the average citations for females is lower than those of the linear and stumbling cluster, which may be explained by the previous observations in \cref{sec:profile_stats} that top percentile citations in all AI scholars are higher than those in female scholars, and this phenomenon might be more pronounced in the exponential cluster.

The smallest cluster is the struggling cluster, where the scholars experience fluctuations in citations but no clear pattern of overall increase. This cluster is also the one that correlates with the highest ratio of scholars that stop publishing, with a percentage of 44+\% for both average scholars and female scholars, followed by the stumbling cluster with 22+\% percentage of people who stop publishing. All these percentages are substantially higher than the stop-publishing percentage in the exponential and linear clusters.



\paragraph{Female Subgroups in the Clusters.}
For each cluster that we identify, we also show the female ratio of each feature for the clusters in \cref{tab:clusters_gender_gap}. Across all the clusters, female researchers are always less than 1/5 of the population, and experience
fewer citations (e.g., with an All:F ratio being 1.58:1 in the exponential cluster, and 1.80:1 in the struggling cluster), and lower h-indices, which are correlated with citaions.
In the stumbling cluster, we see that female reseachers are 1.47\% more likely to stop publishing.

\ifarxiv
As an additional note, we also conduct additional analysis to focus on female subgroups in NLP, and find that female researchers in NLP have higher citations than average female scholars, which is a trend across all clusters.
We include detailed results in the Appendix \cref{tab:cluster_nlp}.
\fi




\section{Analysis of Co-Authorship Patterns}

We also address co-authorship patterns, and conduct analyses to answer the following questions: (1) Do female scholars tend to have more diverse collaborators? And (2) Are there certain gender patterns in different author roles, and what does that indicate for mentor-mentee relationship?


\myparagraph{Aggregated Coauthor Statistics.
}
We are interested in the question ``Does diversity attract diversity?'' A potential angle to understand this is to compare the characteristics of all AI scholars' coauthors and female scholars' coauthors.

\begin{table}[t]
    \centering \small
    \setlength\tabcolsep{1.5pt}
    \resizebox{\columnwidth}{!}{%
    \begin{tabular}{lcc}
    \toprule
         & All & F. \\
    \midrule
    F. Coauthors \% ($\uparrow$)&
    9.58{\tiny $\pm$14.27} &  14.53{\tiny $\pm$26.89} \\
    Coauthors' Domain Diversity ($\uparrow$) &  2.48{\tiny $\pm$0.99} &  2.44{\tiny $\pm$0.96}\\
    \% Coauthors in Freq. Ten Orgs. ($\downarrow$) &  6.76{\tiny $\pm$14.15} &  6.98{\tiny $\pm$14.43} \\
    \% Coauthors in AI Scholars ($\downarrow$) & 18.28{\tiny $\pm$14.39} &  19.85{\tiny $\pm$14.80} \\
    \bottomrule
    \end{tabular}
    }
    \caption{Diversity indices among the coauthors of general AI scholars and female scholars. We use $\uparrow$ and $\downarrow$ to indicate a higher or lower number in this indicator might represent more diversity. Implementation details are in \cref{appd:impl_coauthor_diversity}.}
    \label{tab:coauthor_diversity}
\end{table}

As we can see in \cref{tab:coauthor_diversity}, female scholars have a much larger percentage of female coauthors (14.53\%), which is +4.95\% by absolute value higher than that of the average AI scholars.
Female researchers' other diversity indices are slightly lower,
which might correlate with the previous finding that female scholars are
more concentrated as the ten most frequent organizations. The dynamics of collaboration could be worth exploration in future studies.

\myparagraph{Author Lists of Papers.}
Furthermore, we calculate the statistics based on the author lists of the
    AI papers in the \data{} dataset \citet{jin2022ai}.
In \cref{tab:paper_author_relation}, we investigate that, given the last author's gender, what are some gender patterns in the first author role, and the majority of the authors.

There are some noteworthy conditional probabilities showing large gender disparity. For example, among all papers with male last authors,
there are some astoundingly strong gender disparities -- 1:4.61 female-to-male ratio in the first authors, and 1:48.47 female-to-male ratio for the gender of the majority authors.
Among female last authors, the first author role reaches more gender balance, and the gender ratio of majority authors are reversed, with almost two times more female-majority papers than male-majority papers.
This echoes with the gender homophily observation in \citet{schluter-2018-glass}, although the previous study only focuses on the NLP domain.

\begin{table}[t]
    \centering
    \small
    \begin{tabular}{lcccccc}
    \toprule
    & 1st Author F:M & Majority Authors F:M \\
    \midrule

    Last F. & 1:1.75 & 1:0.54 { }
    \\
    Last M. & 1:4.61 & 1:48.47 \\
    \bottomrule
    \end{tabular}
    \caption{Given the last author's gender, we show the female-to-male gender ratio (F:M) in the first author role (1st Author) and majority authors (i.e.,  >50\% of the authors). We calculate the statistics based on the author lists of the
    AI papers in the \data{} dataset \citet{jin2022ai}.
    }
    \label{tab:paper_author_relation}
\end{table}

Moreover, previous papers have suggested using the relationship between the first author and the last author as a proxy for mentee-mentor relation \cite{schluter-2018-glass}.
From \cref{tab:paper_author_relation}, it seems that male mentors (using the last author as a proxy) tend to take more male mentees (using the first author as a proxy), while female mentors are more balanced, although the ratio is still not equal, perhaps limited by the disparity in the sheer amount of female researchers in the AI community.

\section{Analysis of Female-Authored Papers}

\begin{table}[ht]
    \centering \small
    \setlength\tabcolsep{3pt}
    \begin{tabular}{l@{\extracolsep{\fill}}*{6}{c}}
    \toprule
    & \multicolumn{2}{c}{Citations} & \multicolumn{2}{c}{\# Coauthors} \\ \cmidrule(lr){2-3} \cmidrule(lr){4-5}
    & Avg & $>$95th  & Avg & $>$95th \\
    &
    & (4,599 Papers) &
    & (5,615 Papers) \\
    \midrule
    1st F. & 33 & 305 (6.63\%) & 5 & 468 (8.33\%)\\
    1st M. & 42 & 1,337 (29.07\%) & 5 & 1,147 (20.43\%)\\ \hline
    >50\% F. & 29 & 285 (6.20\%) & 8 & 601 (10.70\%)\\
    >50\% M. & 42 & 2,309 (50.20\%) & 6 & 2,852 (50.79\%)\\ \hline
    Last F. & 32 & 233 (5.07\%) & 5 & 304 (5.41\%) \\
    Last M. & 42 & 1,425 (30.98\%) & 4 & 1,093 (19.47\%)\\
    \bottomrule
    \end{tabular}
    \caption{
    For each group of papers with a certain author gender information, we calculate the average citations and the average number of coauthors. In addition, we also check each coauthor's gender group's presence in the 95th percentile of paper citations (i.e., $>$113 citations) and the number of coauthors (i.e., $>$10 coauthors).
    }
    \label{tab:paper_type_female_rate}
\end{table}

Since our study features a comprehensive bottom-up analysis of the female subgroup, we have covered statistics related to individual scholars and coauthorship, and, finally, in this section, we analyze statistics of female-authored papers.

\subsection{General Paper Statistics}

We first calculate some general statistics of papers with different author gender information in \cref{tab:paper_type_female_rate}. We can see that papers by female authors tend to have more coauthors, while
male authors tend to have on average higher citations than those by female authors at the same authorship position or majority representation, for example, 13 more citations on average for male-majority papers than female-majority papers.
Moreover, this disparity is very large if we zoom into the top papers. Specifically, we take papers over the 95th percentile (with over 113 citations), for example, 2.3K of these papers have a male-majority author list, in contrast to the 285 papers with a female-majority author list (8:1).




\subsection{Linguistic Features of Titles and Abstracts}

Next, we take into consideration the titles and abstracts of all the papers, and calculate their linguistic features.

\paragraph{Frequencies of Different Word Categories.}
We first look at the word categories and their frequency by the Linguistic Inquiry and Word Count (LIWC) 2015 \cite{pennebaker2001linguistic}. We show in \cref{tab:significant_liwc1} a selection of features on which female first-authored papers show a clear difference from average papers, and the comprehensive list of all features on all scholars, female first-author papers, female-majority papers, and female last-authored papers are in \cref{appd:liwc}.
\begin{table}[t]
    \centering
    \setlength\tabcolsep{2pt}
    \small
    \resizebox{\columnwidth}{!}{\begin{tabular}{lclcccc}
    \toprule
    {LIWC Category} \& Top 5 Freq Words & All & 1st F. \\
    \midrule
    Positive Emotion & \multirow{2}{*}{1.98}  &  \multirow{2}{*}{2.05 ({\color{LimeGreen}$\uparrow$3.71\%})}  \\
    \{well, important, energy, better, support\} \\  \hline
    {Female References}  & \multirow{2}{*}{0.01}  &  \multirow{2}{*}{0.02} \multirow{2}{*}{({\color{OliveGreen}$\uparrow$44.48\%})} \\
    \{female, her, women, females, she\} \\ \hline
    {Achievement} & \multirow{2}{*}{2.15}  &  \multirow{2}{*}{2.19 ({\color{LimeGreen}$\uparrow$1.92\%})} \\
    \{first, work, efficient, obtained, better\} \\ \hline
    {Certainty}  & \multirow{2}{*}{0.88}  &  \multirow{2}{*}{0.91 ({\color{LimeGreen}$\uparrow$3.73\%})}
    \\
    \{all, accuracy, specific, accurate, total\}\\ \hline
    {Interrogatives} & \multirow{2}{*}{0.88}  &  \multirow{2}{*}{0.91 ({\color{LimeGreen}$\uparrow$3.44\%})}
    \\
    \{which, when, where, how, whether\} \\ \hline
    {Past Focus}  & \multirow{2}{*}{1.96}  &  \multirow{2}{*}{2.14 ({\color{ForestGreen}$\uparrow$9.34\%})}
    \\
    \{used, was, been, were, obtained\}  \\ \hline
    Present Focus & \multirow{2}{*}{6.29}  &  \multirow{2}{*}{6.26 ({\color{WildStrawberry}$\downarrow$0.56\%})}
    \\
    \{is, are, be, can, have\}  \\
    \bottomrule
    \end{tabular}
    }
    \caption{Linguistic features extracted by LIWC have the most difference between female scholars and male scholars. Each number means occurrence per string (which is abstract). The number in the parentheses shows the relative difference. We also show the top 5 words from score-All.
    We compare features of general abstracts (using the 83K random sample), and features of abstracts of female-authored papers. See std, full word category, etc. in the appendix.}
    \label{tab:significant_liwc1}
\end{table}


In \cref{tab:significant_liwc1}, for example, female first authors tend to use more words about positive emotion, such as ``better'' and ``support.''
Moreover, we can see that female first authors usually use more female references in their papers, which might be due to more female researchers publishing gender-related papers.

We also find it very interesting that the interrogative words are more dominant in female-authored papers, which is probably explained by the writing style difference that female first authors tend to use longer sentences and more commas in their writing (a more detailed analysis of which can be seen in the next paragraph), which might indicate the use of more clauses.
Another interesting fact is that past focus words are more used by female first authors, whereas male first authors' time orientation is a more present focus, which we believe explains another aspect of the writing style difference.

\paragraph{Comprehensive List of Features.}
Apart from the word categories, we also calculate general writing features in \cref{tab:linguistic}. Some distinct features include that female first-author papers tend to have more words in the titles, less use of acronyms, but still more catchy titles. And in the abstract, female first-author papers have more sentences, a larger vocabulary, and more words, while male first-author papers are simpler according to the Flesch readability score \cite{fleschscore}. Also, female first-author papers tend to include numbers more frequently.
\begin{table}[h]
    \centering\small
    \setlength\tabcolsep{1pt}
    \begin{tabular}{lccc}
\toprule
\textbf{Feature} & \textbf{F. } & \textbf{M.} & 
\\
\midrule
\multicolumn{2}{l}{\textbf{\textit{Title Features}}} \\
\quad {\# Words} & 10.08{\tiny $\pm$4.18} & 9.44{\tiny $\pm$4.12} & 
\\
\quad Has Acronym & 3.78\% & 3.86\%& 
\\
\quad {Catchy titles} & 14.96\% & 14.17\% & 
\\
\midrule
\multicolumn{2}{l}{\textbf{\textit{Abstract Features}}} \\
\quad {\# Sentences} & 7.32{\tiny $\pm$4.24} & 6.97{\tiny $\pm$4.18} & 
\\
\quad {\# Vocabulary} & 105.27{\tiny $\pm$43.22} & 101.41{\tiny $\pm$45.75} & 
\\
\quad {\# Words} & 160.46{\tiny $\pm$82.70} & 153.24{\tiny $\pm$79.89} &
\\
\quad TTR & 0.64{\tiny $\pm$0.07} & 0.63{\tiny $\pm$0.08} & 
\\
\quad MATTR density & 91.12{\tiny $\pm$5.37} & 90.97{\tiny $\pm$6.44}  & 
\\
\quad Comma count & 22.57{\tiny $\pm$15.82} & 21.15{\tiny $\pm$14.66} & 
\\
\quad {Flesch Readability} ($\uparrow$) & 10.25{\tiny $\pm$22.72} & 10.94{\tiny $\pm$30.02} &
\\
\quad \# Syllables/Word & 2.02{\tiny $\pm$0.16} & 2.01{\tiny $\pm$0.21} & 
\\
\quad Difficult Word Ratio & 0.30{\tiny $\pm$0.06} & 0.30{\tiny $\pm$0.06} & 
\\
\quad Passive Speech ($\downarrow$) & 0.64{\tiny $\pm$0.47} & 0.64{\tiny $\pm$0.47} &
\\
\quad Uncertainty Tone & 4.78{\tiny $\pm$0.21} & 4.79{\tiny $\pm$0.20} & 
\\
\midrule
\multicolumn{2}{l}{\textbf{\textit{Abstract Content}}} \\
\quad Available on GitHub & 0.49{\tiny $\pm$0.10} & 0.49{\tiny $\pm$0.10} & 
\\
\quad Proposed a Dataset & 0.49{\tiny $\pm$0.12} & 0.49{\tiny $\pm$0.12} & 
\\
\quad Proposed a Task & 0.53{\tiny $\pm$0.11} & 0.53{\tiny $\pm$0.11} & 
\\
\quad SOTA Results & 0.60{\tiny $\pm$0.09} & 0.60{\tiny $\pm$0.09} & 
\\
\quad Has Numbers & 50.34\% & 47.98\% & 
\\
\quad Has Questions & 1.74\%& 1.77\% & 
\\
\bottomrule
    \end{tabular}
    \caption{Linguistic features of papers with female first authors, male first authors, and all. See implementation details in \cref{appd:linguistic}.}
    \label{tab:linguistic}
\end{table}



\paragraph{A Case Study of Title Styles.}
We introduce in detail our identification of catchy titles and findings.
We consider a standard, straightforward paper title as mostly a declarative expression that contains the name of the task and the name of the methodology, while a catchy title is more riveting or humorous, which may involve more diverse forms including questions, quotations, exclamations, and others.
According to these motivations, we build a set of linguistic rules to identify titles that carry catchy styles.
The detailed algorithm is in \cref{appd:stylish_title}. On our self-annotated test set of 1,000 paper titles randomly sampled from the \data{} dataset, our binary classification algorithm achieves 86.3\% F1, with 81.1\% precision and 92.3\% recall, which is significantly higher than the direct application of general catchy website title detection \cite{Mathur_2020} with only 13.2\% F1 scores on our test set.

Apart from the overall observation in \cref{tab:linguistic} that female first-author papers have more catchy titles in general, we can also see from the example titles in \cref{tab:example_titles} that even among catchy titles, male and female authors tend to have different nature of attractiveness in titles, perhaps more creativity, vividness, and humor, at least from a rough glance in our data. A fine-grained analysis could be interesting for future work.


\begin{table}[t]
    \centering \small
    \begin{tabular}{p{7cm}ll}
    \toprule
\multicolumn{1}{c}{\textbf{Example Titles from Male First-Authored Papers}} \\
- Information Power Grid: The new frontier in parallel computing?
\newline
- A systematic review of solid-pseudopapillary neoplasms: Are these rare lesions?
\newline
- Dengue fever again in Pakistan: Are we going in the right direction
 \\ \midrule
\multicolumn{1}{c}{\textbf{Example Titles from Female First-Authored Papers}} \\
- ``I want to slay that Dragon!'' -- Influencing choice in interactive storytelling
    \newline
- Biting off more than we could chew -- A surprising find on biopsy!
\newline
- ‘Spam, Spam, Spam, Spam\dots Lovely Spam!’ Why Is Bluespam Different?
\\
    \bottomrule
    \end{tabular}
    \caption{Stylish titles selected from paper titles of the top 5 scholars that have the largest number of stylish titles among male scholars and female scholars.
    }
    \label{tab:example_titles}
\end{table}

\section{Conclusion}
In this work, we investigated the gender differences in the AI publication world from a comprehensive range of perspectives: basic scholar profile statistics, citation trends, coauthorship, and linguistic styles of papers. We identified that the female subgroup overall still shows underrepresentation and disadvantages in the AI community. However, there are also distinct characteristics of the female subgroup that makes it unique from the general population. Our analysis provides a window to look at the current trends in our AI community, and encourages more gender equality and diversity in the future.

\section*{Limitations}\label{sec:limitations}
It is very challenging to conduct such a large-scale and diverse-view study on gender differences in the AI publication world. Our limitations are mainly from three perspectives: the inherent difficulties of identifying each term, inevitable noises in the data, and our method which is mainly correlational analysis.

The inherent difficulties of identifying each term are the largest limitation and constraint for this type of study, starting from the difficult process to decide some ethically-sensitive terms such as gender based on balancing concerns over both feasibility and ethics, to deciding some human-interpretable but slightly subjective categories such as catchy titles, which is a balance over reader-friendliness of the results and objectivity of the feature identification.

Another challenge is the inevitable noises in the data. Also, the most important noises come from the identification of gender, where we have to stay relatively conservative and leave a large portion of the author genders undecided, not to mention the errors for researchers whose names can be matched with names in the database but the self-identification of gender could still vary case by case. Apart from this, there are also various other noises such as selection biases. For example, not all AI researchers establish a Google Scholar profile or tag themselves in the AI domain on their profile, among many other sources of noise.

The third limitation is that our study is mainly based on analysis over correlations. It is not suggestive to directly use the study to guide interventions or decision-making, since our conclusions have not nailed down to causal factors of the disparities related to gender. In future work, it is very welcome to investigate more and use causal inference to identify, for example, mediators of academic success that provide equal opportunities for all genders.







\section*{Ethical Considerations}
The ethical considerations of this study mainly overlap with our limitations. There is no perfect way when it comes to conclusions related to gender. We deeply understand that gender is highly personal and diverse in nature. In this study, we have to take a difficult step to balance the practical and ethical concerns, since the large-scale statistical analysis needs to be based on the identification of gender in a relatively scalable way. We do not wish to harm anyone, while in the meantime we try to bring as informative analysis that could be helpful for the community to understand the underrepresentation of the female subgroup on various axes as possible. We are very welcome for follow-up discussions on the ethics of this study, and we are open to improvements accordingly.

\ifarxiv
\section*{Acknowledgment}

We appreciate constructive discussions with Saif M. Mohammad (Senior Research Scientist at National Research Council Canada) and his generous sharing of previous experience in his NLP Scholar project.

This material is based in part upon works supported by the German Federal Ministry of Education and Research (BMBF): Tübingen AI Center, FKZ: 01IS18039B; by the Machine Learning Cluster of Excellence, EXC number 2064/1 – Project number 390727645; by the Precision Health Initiative at the University of Michigan; by the John Templeton Foundation (grant \#61156); and by a Responsible AI grant by the Haslerstiftung.

\fi

\bibliography{sec/ref_anthology,custom,sec/ref_ai_scholar,sec/ref_causality,sec/refs_zhijing}

\begin{thebibliography}{18}
\expandafter\ifx\csname natexlab\endcsname\relax\def\natexlab#1{#1}\fi

\bibitem[{Birhane et~al.(2022)Birhane, Kalluri, Card, Agnew, Dotan, and
  Bao}]{birhane2021values}
Abeba Birhane, Pratyusha Kalluri, Dallas Card, William Agnew, Ravit Dotan, and
  Michelle Bao. 2022.
\newblock \href {https://facctconference.org/static/pdfs_2022/facct22-14.pdf}
  {The values encoded in machine learning research}.

\bibitem[{Deng et~al.(2009)Deng, Dong, Socher, Li, Li, and
  Fei{-}Fei}]{deng2009imagenet}
Jia Deng, Wei Dong, Richard Socher, Li{-}Jia Li, Kai Li, and Li~Fei{-}Fei.
  2009.
\newblock \href {https://doi.org/10.1109/CVPR.2009.5206848} {Imagenet: {A}
  large-scale hierarchical image database}.
\newblock In \emph{2009 {IEEE} Computer Society Conference on Computer Vision
  and Pattern Recognition {(CVPR} 2009), 20-25 June 2009, Miami, Florida,
  {USA}}, pages 248--255. {IEEE} Computer Society.

\bibitem[{Hand et~al.(2017)Hand, Rice, and Greenlee}]{hand2017exploring}
Sarah Hand, Lindsay Rice, and Eric Greenlee. 2017.
\newblock Exploring teachers’ and students’ gender role bias and
  students’ confidence in stem fields.
\newblock \emph{Social Psychology of Education}, 20(4):929--945.

\bibitem[{Honnibal and Montani(2017)}]{spacy2}
Matthew Honnibal and Ines Montani. 2017.
\newblock {spaCy 2}: Natural language understanding with {B}loom embeddings,
  convolutional neural networks and incremental parsing.
\newblock To appear.

\bibitem[{Jin et~al.(2022)Jin, Lyu, Ding, Sachan, Zhang, Mihalcea, and
  Schoelkopf}]{jin2022ai}
Zhijing Jin, Zhiheng Lyu, Yiwen Ding, Mrinmaya Sachan, Kun Zhang, Rada
  Mihalcea, and Bernhard Schoelkopf. 2022.
\newblock \href {https://zhijing-jin.com/files/papers/AIScholar_2022.pdf} {{AI
  Scholars: A} dataset for {NLP}-involved causal inference}.

\bibitem[{Krizhevsky et~al.(2012)Krizhevsky, Sutskever, and
  Hinton}]{krizhevsky2012imagenet}
Alex Krizhevsky, Ilya Sutskever, and Geoffrey~E. Hinton. 2012.
\newblock \href
  {https://proceedings.neurips.cc/paper/2012/hash/c399862d3b9d6b76c8436e924a68c45b-Abstract.html}
  {{ImageNet} classification with deep convolutional neural networks}.
\newblock In \emph{Advances in Neural Information Processing Systems 25: 26th
  Annual Conference on Neural Information Processing Systems 2012. Proceedings
  of a meeting held December 3-6, 2012, Lake Tahoe, Nevada, United States},
  pages 1106--1114.

\bibitem[{Mathur(2020)}]{Mathur_2020}
Saurabh Mathur. 2020.
\newblock \href {https://doi.org/10.17605/OSF.IO/T3UJ9} {clickbait-detector}.

\bibitem[{Mohammad(2020)}]{mohammad-2020-gender}
Saif~M. Mohammad. 2020.
\newblock \href {https://doi.org/10.18653/v1/2020.acl-main.702} {Gender gap in
  natural language processing research: Disparities in authorship and
  citations}.
\newblock In \emph{Proceedings of the 58th Annual Meeting of the Association
  for Computational Linguistics}, pages 7860--7870, Online. Association for
  Computational Linguistics.

\bibitem[{Pennebaker et~al.(2001)Pennebaker, Francis, and
  Booth}]{pennebaker2001linguistic}
James~W Pennebaker, Martha~E Francis, and Roger~J Booth. 2001.
\newblock Linguistic inquiry and word count: Liwc 2001.
\newblock \emph{Mahway: Lawrence Erlbaum Associates}, 71(2001):2001.

\bibitem[{Reimers and Gurevych(2019)}]{reimers-sentence-bert}
Nils Reimers and Iryna Gurevych. 2019.
\newblock \href {https://arxiv.org/abs/1908.10084} {Sentence-bert: Sentence
  embeddings using siamese bert-networks}.
\newblock In \emph{Proceedings of the 2019 Conference on Empirical Methods in
  Natural Language Processing}. Association for Computational Linguistics.

\bibitem[{Ritchie and Roser(2019)}]{owidgenderratio}
Hannah Ritchie and Max Roser. 2019.
\newblock Gender ratio.
\newblock \emph{Our World in Data}.
\newblock Https://ourworldindata.org/gender-ratio.

\bibitem[{Robnett(2016)}]{robnett2016gender}
Rachael~D Robnett. 2016.
\newblock Gender bias in stem fields: Variation in prevalence and links to stem
  self-concept.
\newblock \emph{Psychology of women quarterly}, 40(1):65--79.

\bibitem[{Schluter(2018)}]{schluter-2018-glass}
Natalie Schluter. 2018.
\newblock \href {https://doi.org/10.18653/v1/D18-1301} {The glass ceiling in
  {NLP}}.
\newblock In \emph{Proceedings of the 2018 Conference on Empirical Methods in
  Natural Language Processing}, pages 2793--2798, Brussels, Belgium.
  Association for Computational Linguistics.

\bibitem[{Stathoulopoulos and Mateos-Garcia(2019)}]{stathoulopoulos2019gender}
Konstantinos Stathoulopoulos and Juan~C Mateos-Garcia. 2019.
\newblock Gender diversity in ai research.
\newblock \emph{Available at SSRN 3428240}.

\bibitem[{Talburt(1986)}]{fleschscore}
John Talburt. 1986.
\newblock \href {https://doi.org/10.1145/10563.10583} {The flesch index: An
  easily programmable readability analysis algorithm}.
\newblock In \emph{Proceedings of the 4th Annual International Conference on
  Systems Documentation}, SIGDOC '85, page 114–122, New York, NY, USA.
  Association for Computing Machinery.

\bibitem[{Tanveer et~al.(2018)Tanveer, Samrose, Baten, and
  Hoque}]{Tanveer-2018}
M.~Iftekhar Tanveer, Samiha Samrose, Raiyan~Abdul Baten, and M.~Ehsan Hoque.
  2018.
\newblock \href {https://doi.org/10.1145/3173574.3173598} {Awe the audience:
  How the narrative trajectories affect audience perception in public
  speaking}.
\newblock In \emph{Proceedings of the 2018 CHI Conference on Human Factors in
  Computing Systems}, CHI '18, page 1–12, New York, NY, USA. Association for
  Computing Machinery.

\bibitem[{Tavenard et~al.(2020)Tavenard, Faouzi, Vandewiele, Divo, Androz,
  Holtz, Payne, Yurchak, Ru{\ss}wurm, Kolar, and Woods}]{JMLR:v21:20-091}
Romain Tavenard, Johann Faouzi, Gilles Vandewiele, Felix Divo, Guillaume
  Androz, Chester Holtz, Marie Payne, Roman Yurchak, Marc Ru{\ss}wurm, Kushal
  Kolar, and Eli Woods. 2020.
\newblock \href {http://jmlr.org/papers/v21/20-091.html} {Tslearn, a machine
  learning toolkit for time series data}.
\newblock \emph{Journal of Machine Learning Research}, 21(118):1--6.

\bibitem[{Vogel and Jurafsky(2012)}]{vogel-jurafsky-2012-said}
Adam Vogel and Dan Jurafsky. 2012.
\newblock \href {https://aclanthology.org/W12-3204} {He said, she said: Gender
  in the {ACL} {A}nthology}.
\newblock In \emph{Proceedings of the {ACL}-2012 Special Workshop on
  Rediscovering 50 Years of Discoveries}, pages 33--41, Jeju Island, Korea.
  Association for Computational Linguistics.

\end{thebibliography}
\bibliographystyle{acl_natbib}

\clearpage
\newpage

\appendix

\section{Domain Tag Cleaning} \label{appd:domain_tag_cleaning}
Among 30,596 unique domains all scholars have, we manually extract 26 general domains and merge them with their sub-domains. To account for the fact that some scholars might label themselves with more than one domain, we normalize the count by 1 / the number of domains they identify themselves with.

\paragraph{Limitations of Self-Labeling:}
We acknowledge that some domains have fewer samples, which may lead to a deviation in the female percentage. However, it should be noted that the female percentage in NLP, AI, and CV and their ranking in \cref{fig:domain} are matched with the result in \cref{tab:female_rate}.

\section{Analysis of the Population with Unclassified Gender}\label{appd:unclassified}
\begin{figure}[h]
\centering
  \minipage{\columnwidth}
    \centering
    \includegraphics[width=0.8\columnwidth]{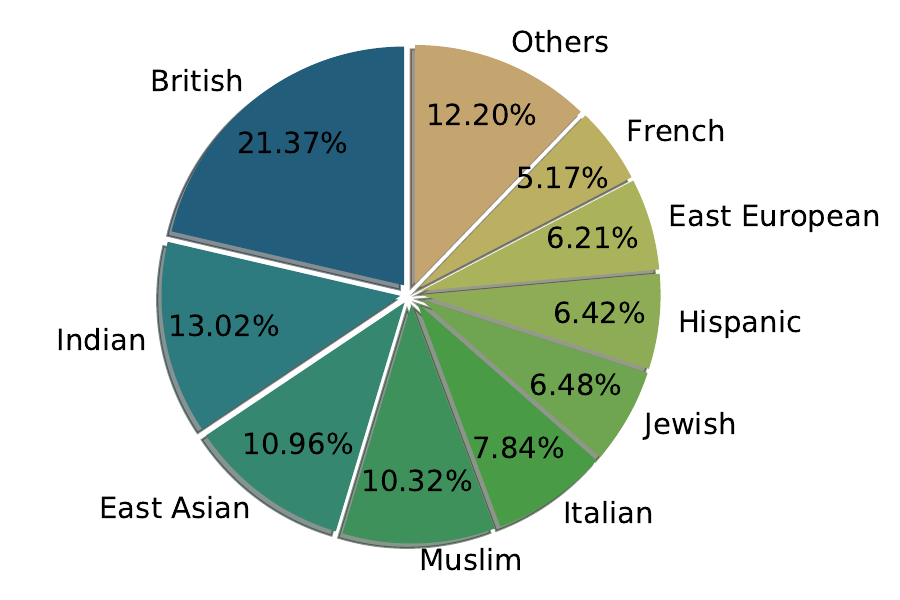}
  \endminipage
  \hfill
  \minipage{\columnwidth}
    \centering
    \includegraphics[width=0.8\columnwidth]{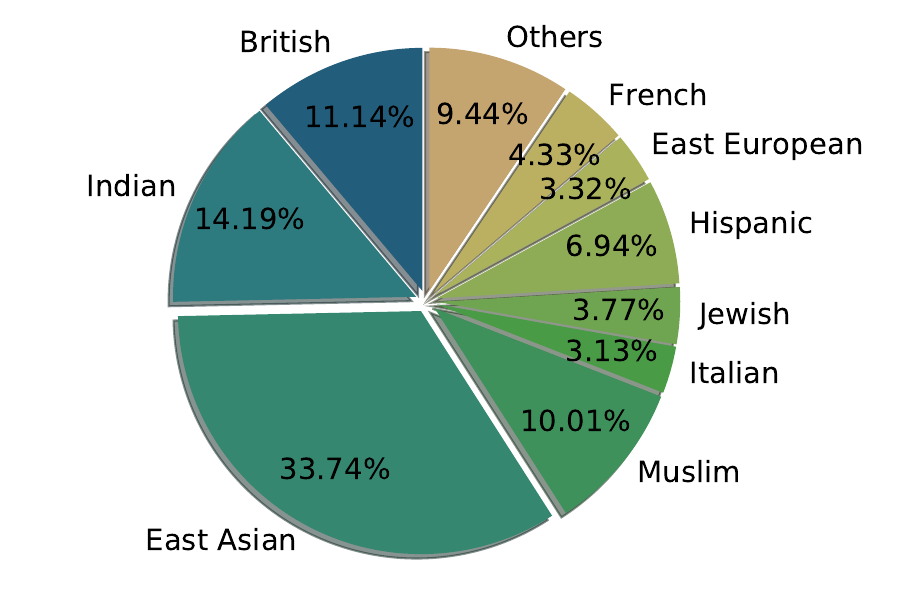}
  \endminipage
    \caption{Ethnicity distributions of classified (left) and unclassified (right) researchers.}
    \label{fig:ethnicity_unclassified}
\end{figure}

Our data is inclusive for various ethinicities, as shown in the left subfigure of \cref{fig:ethnicity_unclassified}. However, our dataset subsamples certain groups such as east Asians, Indians and so on, as in the right subfigure of \cref{fig:ethnicity_unclassified}. Some cases might be intractable. For example, Chinese names have gender markers only in their own writing system, so the gender markers are lost if we use the Romanized spelling of Chinese names on Google Scholar, leaving it only possible to classify the gender using researchers' photos, which is unethical.

\begin{figure}[h]
  \centering
  \minipage{\columnwidth}
    \centering
    \includegraphics[width=\columnwidth]{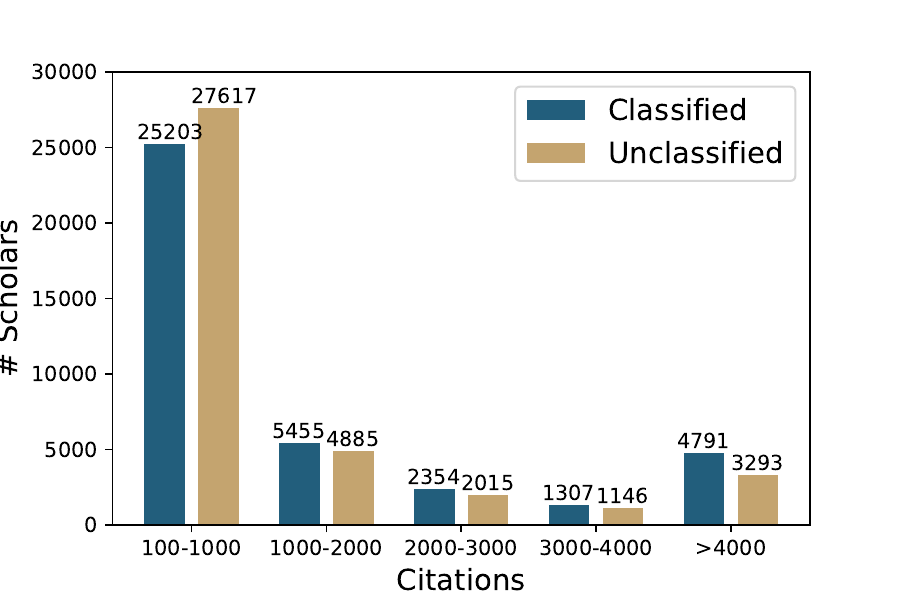}
  \endminipage
  \hfill
  \minipage{\columnwidth}
    \centering
    \includegraphics[width=\columnwidth]{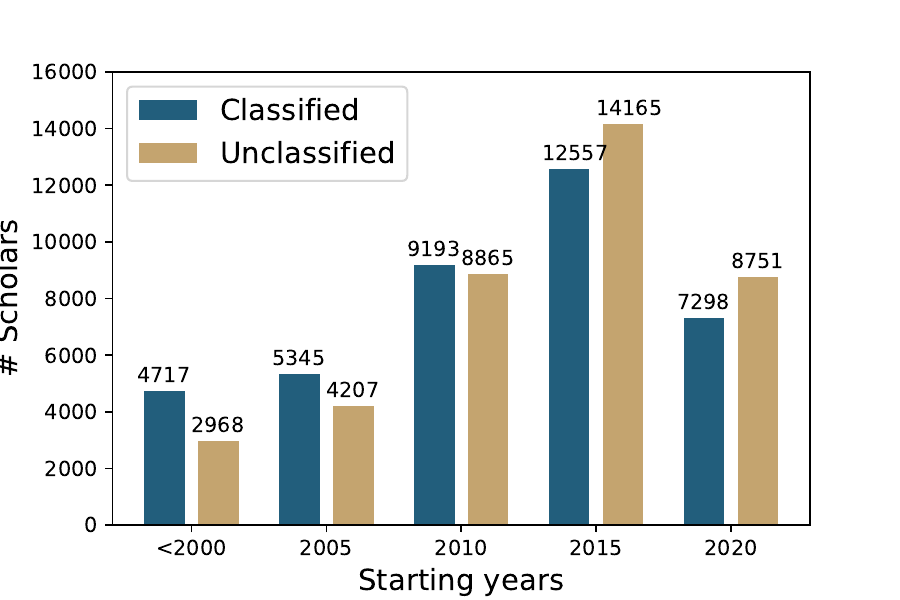}
  \endminipage
    \caption{Left: Citation distributions of the researcher population with classified and unclassified gender. Right: Histogram of the publication starting year (i.e., the year of the first paper according to Google Scholar) of the researcher population with classified and unclassified gender.}
    \label{fig:profile_unclassified}
\end{figure}

Further, the left-out population carry overall similar characteristics with our gender-identified population. From the analysis of profile statistics
of unclassified researchers in \cref{fig:profile_unclassified},
we can see that the citation distribution and starting years of the two population are roughly similar, with the unclassified population is slightly younger and thus less cited.


\section{General Profile Statistics}\label{appd:general_profile}
We calculate more statistics of AI researchers' profiles. \cref{tab:profile_num_features_std} shows standard deviations of the features in \cref{tab:profile_num_features}. \cref{tab:cit_by_2012} includes the statistics of citation within different year spans, from which we can see that female scholars' citation is generally less than all scholars'.  \cref{tab:female_rate_by_aca_age} confirms that female scholars take a higher percentage in younger academic age groups than in senior groups.

\begin{table}[ht]
    \centering
    \small
    \setlength\tabcolsep{2pt}
    \resizebox{\columnwidth}{!}{     \begin{tabular}{p{1.5cm}lccccccc}
    \toprule
    & & Avg (std) & Min & 25th & 50th & 75th & Max \\
    \midrule
    Citations: & All & 2,129.54{\tiny $\pm$8,639.88} & 100 & 214 & 475 & 1,345 & 533,757\\
    & F. & 1,762.11{\tiny $\pm$6,246.01} & 100 & 197 & 414 & 1,165 & 209,549 \\ \hline
    h-Index: & All & 14.03{\tiny $\pm$13.24} & 1 & 7 & 10 & 16 & 266\\
    & F. & 13.25{\tiny $\pm$12.67} & 1 & 6 & 9 & 15 & 211 \\ \hline
    \# Papers: & All & 67.44{\tiny $\pm$127.00} & 1 & 17 & 32 & 68 & 3,000\\
    & F. & 60.20{\tiny $\pm$103.96} & 1 & 16 & 29 & 64 & 2,125\\ \hline
    AcadAge: & All & 16.89{\tiny $\pm$10.65} & 2 & 10 & 14 & 20 & 72\\
    & F. & 16.33{\tiny $\pm$9.74} & 2 & 10 & 14 & 20 & 73\\ \hline
    Active Yrs.: & All & 15.47{\tiny $\pm$10.75} & 1 & 8 & 13 & 19 & 72\\
    & F. & 14.87{\tiny $\pm$9.88} & 1 & 8 & 12 & 19 & 72\\ \hline
    F. Coauthor & All & 9.58{\tiny $\pm$ 6.18} &0 &5.21 & 8.70 & 12.50 & 48.72  \\
    (\%): & F. & 14.53{\tiny $\pm$ 7.52} & 1.64 & 9.09 & 13.04 & 18.58 & 48.72 \\
    \bottomrule
    \end{tabular}
    }
    \caption{A more comprehensive version of \cref{tab:profile_num_features} including the standard deviation.}
    \label{tab:profile_num_features_std}
\end{table}

\begin{table}[ht]
    \centering
    \small
    \setlength\tabcolsep{2pt}
    \resizebox{\columnwidth}{!}{
    \begin{tabular}{p{1.5cm}ccccccc}
    \toprule
    Citations & Avg (std) & Min & 25th & 50th & 75th & Max \\
    \midrule
     < Yr 2012 & 715.97{\tiny$\pm$2841.15} & 1 & 18 & 83 & 371 & 122,289 \\ \hline
     < Yr 2012 (F) & 555.41{\tiny $\pm$1832.53} & 1 & 16 & 75 & 343 & 31,130\\ \hline
     >= Yr 2012 & 2484.98{\tiny $\pm$8908.21} & 0 & 280 & 698 & 1,885 & 467,586\\ \hline
     >= Yr 2012 (F) & 2120.10{\tiny $\pm$6901.53} & 0 & 261 & 639 & 1,670 & 203,008\\ \hline
     >= Yr 2012 (newbies) & 905.14{\tiny $\pm$4504.45} & 2 & 165 & 285 & 621 & 287,603\\ \hline
     >= Yr 2012 (newbies,F) & 734.65{\tiny $\pm$3063.95} & 15 & 156 & 250 & 521 & 79,245\\
    \bottomrule
    \end{tabular}
    }
    \caption{Descriptive statistics of citation breakdown. We compare the citation for the total population with the citation only for female scholars, where female citations are generally fewer.}
    \label{tab:cit_by_2012}
\end{table}

\begin{table}[ht]
    \centering
    \small
    \resizebox{\columnwidth}{!}{     \begin{tabular}{ccccc}
    \toprule
    Academic Age & \# F & \# M & \# All & F Rate \\
    \midrule
    0-5 & 454 & 1,848 & 5,271 & 24.56 \\
    5-10 & 2,242 & 9,136 & 24,458 & 24.54 \\
    10-15 & 1,877 & 8,270 & 20,965 & 22.69 \\
    15-20 & 1,126 & 5,391 & 11,951 & 9.42 \\
    20-25 & 496 & 2,573 & 5,338 & 9.29 \\
    25-30 & 235 & 1,371 & 2,598 & 9.05 \\
    30-35 & 86 & 665 & 1,152 & 12.9 \\
    35-40 & 44 & 394 & 704 & 6.25 \\
    \bottomrule
    \end{tabular}
    }
    \caption{The number of female scholars, male scholars, and total scholars in different groups of academic age. Female scholars take up a much higher proportion in younger academic age groups than in senior groups, while for male scholars the opposite is true.}
    \label{tab:female_rate_by_aca_age}
\end{table}

\section{Implementation Details} \label{implementation}
\subsection{Academia and Orgs}\label{appd:impl_academia_and_orgs}
\paragraph{Identification of Academia Status:}
We define whether a scholar belongs to academia by their description in the GS profile. We use keyword matches such as ``university'', ``professor'' etc., to determine their academic status. If there is no evidence that a scholar is in academia, we will label the scholar in industry.

\paragraph{Extraction Method for Top10 Organizations:}
We use the description of each scholar (e.g., Professor of Computer Science, University of Michigan) in our \data{} to classify their organizations. Google Scholar itself has a unique code for a wide range of organizations, and the discrepancy in position description will not affect the organization code. Thus we first cluster the organizations with unique codes and get 3568 organizations in total. For those without unique code from GS, we use Named Entity Recognition by \citet{spacy2} to filter out the plain organization (ORG) in their description. Then we employ sentence embedding followed by a  fast clustering algorithm \cite{reimers-sentence-bert} with a cosine similarity threshold of 0.75 to cluster the organization, which results in 220 clusters. With the organization results combined from the above two methods, we obtain the top 10 most frequent organizations as Google, Stanford, CMU, MIT, Amazon, UCB, Microsoft, Facebook, IBM, and Apple.


\subsection{Time Series Implementation}\label{appd:cluster_time_trends}

We simplify the method from \citet{Tanveer-2018}, where they first smooth the trajectories by a 5-point average kernel and standardize the trajectories by subtracting the time average and dividing by the time-wise standard deviation. Instead, we interpolate the citation time span to the longest active academic age and normalize the trajectories by their average citation number to focus on the relative rises and falls. In addition, we use TimeSeriesKMeans with DTW metric to cluster the trajectories, instead of density-based clustering (DBSCAN) employed by \citet{Tanveer-2018}, as DTW metric is used specifically to collect time series of similar shapes.
\begin{algorithm}[H]
    \caption{Generating the Time Series Clusters}
    \label{alg:cluster_time_trends}
    \begin{algorithmic}
        \REQUIRE{Citations vs. years for 78k scholars}
        \ENSURE{9 Clusters of relative citations time trends}
        \STATE $\text{max\_num} \leftarrow \max{(\text{len}(\text{years\_list)}} $
        \STATE $\text{cites\_upd\_list} \leftarrow \text{empty list}$
        \FOR{years, cites in zip(years\_list, cites\_list)}
            \STATE $\text{itp\_cites} \leftarrow \text{interp(cites, max\_num)}$
            \STATE $ \text{cites\_upd\_list append itp\_cites}/ \mu\text{(itp\_cites)}$
        \ENDFOR
        \STATE $\text{model} \leftarrow \text{TimeSeriesKMeans(cites\_upd\_list,}$
        \STATE $\text{\quad n\_clusters=9, metric=`dtw')}$
    \end{algorithmic}
\end{algorithm}

\subsection{Coauthor Diversity}\label{appd:impl_coauthor_diversity}

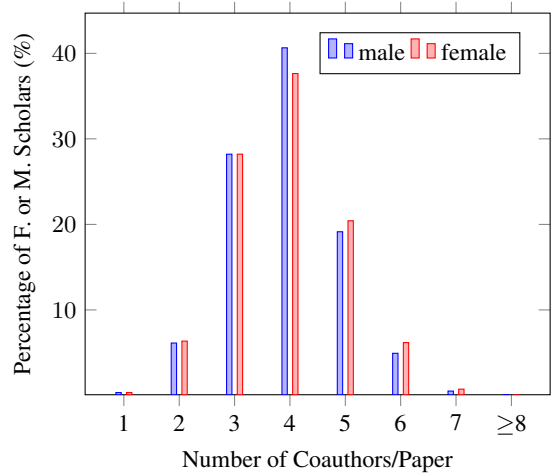
\begin{figure}[t]
\begin{tikzpicture}
\footnotesize
\begin{axis}
[
    ybar, 
    bar width=2pt,
    width=\columnwidth,
    legend style={at={(0.72,0.95)}, 
      anchor=north,legend columns=-1},
    ylabel={Percentage of F. or M. Scholars (\%)}, 
    xlabel={Number of Coauthors/Paper},
    symbolic x coords={{1},{2},{3},{4},{5},{6},{7},{$\geq$8}},
    xtick=data,
    enlarge y limits={upper=0},
    ]
\addplot coordinates {(1, 0.337) (2, 6.130) (3, 28.207) (4, 40.644) (5, 19.140) (6,4.929) (7, 0.514) ($\geq$8, 0.087)}; 
\addplot coordinates {(1, 0.341) (2, 6.353) (3, 28.212) (4, 37.635) (5, 20.424) (6,6.182) (7, 0.739) ($\geq$8, 0.085)};
\legend{male,
female}; 

\end{axis}
\end{tikzpicture}
    \caption{Female, non-female and male scholars coauthorship.
    }
    \label{fig:coauthorship_allcoauthors}
\end{figure}

Without specific clarifications, we use ``all coauthors'' in the main text by default, as in \cref{tab:coauthor_diversity}. We measure this by joining the features of coauthors of all papers that a scholar has. \cref{fig:coauthorship_allcoauthors} plots the percentage of scholars with the different number of coauthors. Female scholars tend to write a paper with slightly more coauthors.

In addition, the group of people, \textbf{scholars' female coauthors} in our dataset, is different from what we mention as \textbf{female scholars}. Although many of our experiments mainly work on female coauthors that are also in our dataset (who we have a detailed analysis of their features), they may or may not be findable in our dataset, depending on whether they listed themselves as in AI fields and whether their citations are over 100.

To get the coauthors' domain diversity of a scholar, we union sets of domain tags for ``all coauthors'' of the scholar, and divide the set size by the number of coauthors.

\section{Additional Basic Stats}
\subsection{General AI Subdomains}
We calculate the female scholar percentage in some main AI domains in \cref{tab:female_rate}. We check the percentage of papers that have a female first author and female last author. The table shows that the computer vision domain has the lowest female percentage whereas the natural language processing domain has the highest female percentage. This trend also extends to the female first author paper and female last author paper in every domain.

\begin{table}[ht]
    \centering
    \small
    \resizebox{\columnwidth}{!}{     \begin{tabular}{lcccc}
    \toprule
    & \% F. Scholars & Paper 1st F. (\%) & Paper Last F. (\%) \\
    \midrule
    \textbf{All} & 17.99 & 21.66 & 16.94\\
    AI & 17.27 & 21.12 & 17.43\\
    CV & 15.57 & 19.66 & 14.30\\
    ML & 17.08 & 21.18 & 16.26\\
    NLP & 24.89 & 27.14 & 23.31\\
    \bottomrule
    \end{tabular}
    }
    \caption{Female author rate in 4 different fields. We calculate the rate by \# females/(\# females + \# males) for papers. Note that the female percentage in NLP is the highest.}
    \label{tab:female_rate}
\end{table}


\subsection{Age Groups}
\begin{figure}[ht]

\begin{tikzpicture}
\footnotesize
    \begin{axis}[
        xlabel=Academic Age,
        ylabel=Citations,
        xmin=0, xmax=40,
        ymin=0, ymax=10000,
        xtick={0,5,10,15,20,25,30,35,40},
        ytick={2000,4000,6000,8000,10000},
        yticklabel style={
        /pgf/number format/fixed,
        },
        scaled y ticks=false,
        width=0.9\columnwidth,
        height=0.7\columnwidth,
        legend style={at={(0.5,0.3)}}
        ]
    \addplot[dashed,
    blue] plot coordinates {
        (0, 13.102515819947223)
        (1, 45.39745807228607)
        (2, 103.8879773327323)
        (3, 187.0222122563115)
        (4, 292.89309119182496)
        (5, 410.4767282683094)
        (6, 539.197276716572)
        (7, 656.5527449346782)
        (8, 803.3432130247156)
        (9, 983.1448975549638)
        (10, 1161.377146994208)
        (11, 1352.0020488887624)
        (12, 1614.5967044486667)
        (13, 1846.2205676707358)
        (14, 2120.1251363021765)
        (15, 2435.752226639159)
        (16, 2772.7797762024725)
        (17, 3123.278437564358)
        (18, 3473.345807289553)
        (19, 3851.9331470956554)
        (20, 4210.141308031176)
        (21, 4577.284237040965)
        (22, 4924.323620582765)
        (23, 5317.477657657658)
        (24, 5412.552653748947)
        (25, 5671.0603236880825)
        (26, 5881.973925501433)
        (27, 6072.0162546562815)
        (28, 6002.266357112706)
        (29, 6368.443246019518)
        (30, 6638.669781931464)
        (31, 7168.319665907366)
        (32, 7795.17962962963)
        (33, 8370.506063947078)
        (34, 7492.556910569106)
        (35, 7874.763912310286)
        (36, 8376.025917926567)
        (37, 8508.098143236075)
        (38, 9405.937704918033)
    };
    \addlegendentry{All Scholars}

    \addplot[smooth,
    color=red]
        plot coordinates {
        (0, 11.577885162023877)
        (1, 42.30405117270789)
        (2, 97.3292805035045)
        (3, 174.03838855641513)
        (4, 269.4008891614288)
        (5, 375.68139993427536)
        (6, 483.82287756925825)
        (7, 589.7034632034632)
        (8, 718.3267716535433)
        (9, 894.008572128337)
        (10, 1068.10183639399)
        (11, 1272.2636450686243)
        (12, 1402.9295154185022)
        (13, 1640.8164291701592)
        (14, 1930.070243902439)
        (15, 2105.8856819468024)
        (16, 2365.0532623169106)
        (17, 2658.8170347003156)
        (18, 2954.4555659494854)
        (19, 3065.4161915621435)
        (20, 3432.6698895027625)
        (21, 3643.745542949757)
        (22, 3853.45179584121)
        (23, 4176.178970917226)
        (24, 4127.341463414634)
        (25, 4317.765079365079)
        (26, 4407.748148148148)
        (27, 4698.886255924171)
        (28, 4731.412121212121)
        (29, 5359.4)
        (30, 5366.5)
        (31, 4658.202531645569)
        (32, 5362.65625)
        (33, 5883.320754716981)
        (34, 6465.8536585365855)
        (35, 6334.636363636364)
        (36, 6783.96)
        (37, 7054.391304347826)
        (38, 6653.222222222223)
        };
    \addlegendentry{F. Scholars}
    \end{axis}
    \end{tikzpicture}
    \caption{Citations by academic age of female scholars and total scholars.
    The plot shows that the gap in total citations between female scholars and total scholars keeps widening, as the academic age increases from 15.
    In \cref{tab:avg_citation_aca_age}, the column ``Male Citation / Female Citation'' shows a similar trend in another angle.
    }
    \label{fig:citation_over_year}
\end{figure}
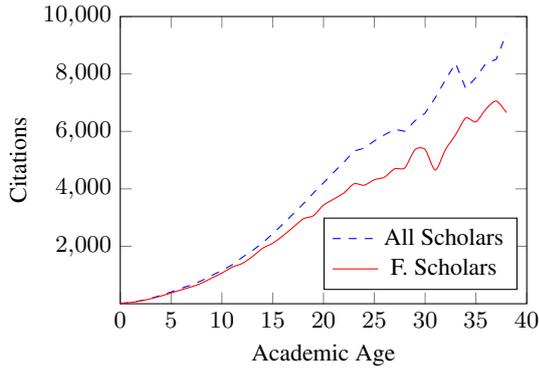

From \cref{fig:citation_over_year}, we can see that the citation difference is not very large at the beginning of the career, but as we proceed to academic age groups of 15, the difference gradually shows up, and becomes larger in more senior academic age groups.

\begin{table}[ht]
    \centering
    \small
    \resizebox{\columnwidth}{!}{
    \begin{tabular}{lcccccccc}
    \toprule
    A. Age & Avg Citations & F. Citations  & M. Citations & M.C:F.C \\ \midrule
    Overall & 2122.80 $\pm$ 8626.45 & 1757.09 $\pm$ 6236.91 & 2694.15 $\pm$ 10921.73 & 1.5333\\
    0 -- 5 & 361.57 $\pm$ 706.41 & 317.38 $\pm$ 413.36 & 385.57 $\pm$ 984.26 & 1.2148\\
    6 -- 10 & 675.11 $\pm$ 2443.80 & 590.68 $\pm$ 2425.48 & 736.15 $\pm$ 2651.44 & 1.2463\\
    11 -- 15 & 1182.22 $\pm$ 5045.14 & 1142.19 $\pm$ 7064.93 & 1398.30 $\pm$ 6042.53 & 1.2242\\
    16 -- 20 & 1997.78 $\pm$ 6098.71 & 1787.87 $\pm$ 4258.76 & 2316.87 $\pm$ 7897.54 & 1.2959\\
    21 -- 25 & 3336.25 $\pm$ 7973.33 & 3483.79 $\pm$ 8469.01 & 3801.13 $\pm$ 9051.56 & 1.0911\\
    $\geq$26 & 7053.99 $\pm$ 18813.80 & 5354.27 $\pm$ 9918.48 & 8218.15 $\pm$ 22185.58 & 1.5350

    \\
    \bottomrule
    \end{tabular}
    }
    \caption{Average citations of all scholars and female scholars in different academic age groups. Female scholars' average citations are less than average citations in nearly all academic ages, and vice versa for male scholars.
    }
    \label{tab:avg_citation_aca_age}
\end{table}

\subsection{Age-Specific Dropout}\label{appd:dropout}

In \cref{tab:domancy_table_with_change}, we show relations between the dropout rate and academic status given different academic age spans. In general, scholars that are in the industry have a much higher dropout rate. In addition, the dropout rate first does up and then down as the academic ages grow, and in academic age 6 - 10 (the time right after Ph.D.), female scholars are less likely to dropout in the industry while male scholars are not.

\begin{table*}[t]
    \centering
    \resizebox{\textwidth}{!}{\begin{tabular}{lccccc}
    \toprule
    Academic Age & F. Dropout (\%) & M. Dropout (\%)& F. Industry (\%) & M. Industry (\%)\\
     & Among All $\rightarrow$ Among Industry & All $\rightarrow$ Industry & All $\rightarrow$ Dropout & All $\rightarrow$ Dropout\\\midrule
    0 -- 5 &  &  & 28.99 $\rightarrow$ 0 & 33.17 $\rightarrow$ 0\\
    6 -- 10
    & 3.10 $\rightarrow$ 2.71 ({\color{WildStrawberry}{$\downarrow$0.39}}) & 3.59 $\rightarrow$ 3.94 ({\color{LimeGreen}{$\uparrow$0.35}})& 38.17 $\rightarrow$ 33.33 (\red{$\downarrow$4.84}) & 42.30 $\rightarrow$ 46.34 ({\color{ForestGreen}{$\uparrow$4.04}})\\
    11 -- 15 & 10.89 $\rightarrow$ 13.07 ({\color{LimeGreen}{$\uparrow$2.18}})& 11.41 $\rightarrow$ 13.36 ({\color{LimeGreen}{$\uparrow$1.95}})& 40.05 $\rightarrow$ 48.08 ({\color{ForestGreen}{$\uparrow$8.03}})& 46.70 $\rightarrow$ 54.68 ({\color{ForestGreen}{$\uparrow$7.98}})\\
    16 -- 20 & 14.95 $\rightarrow$ 21.51 ({\color{ForestGreen}{$\uparrow$6.56}})& 14.07 $\rightarrow$ 17.71 ({\color{LimeGreen}{$\uparrow$3.64}})& 43.58 $\rightarrow$ 62.71 ({\color{OliveGreen}{$\uparrow$19.13}})& 48.64 $\rightarrow$ 61.25 ({\color{ForestGreen}{$\uparrow$12.61}})\\
    21 -- 25 & 8.49 $\rightarrow$ 13.85 ({\color{ForestGreen}{$\uparrow$5.36}})& 13.00 $\rightarrow$ 19.83 ({\color{ForestGreen}{$\uparrow$6.83}})& 35.65 $\rightarrow$ 58.18 ({\color{OliveGreen}{$\uparrow$22.53}})& 45.46 $\rightarrow$ 69.37({\color{OliveGreen}{$\uparrow$23.91}})\\
    26 -- 30 & 10.80 $\rightarrow$18.03 ({\color{ForestGreen}{$\uparrow$7.23}}) & 10.53 $\rightarrow$ 16.99 ({\color{ForestGreen}{$\uparrow$6.46}})& 33.80 $\rightarrow$ 56.41 ({\color{OliveGreen}{$\uparrow$22.61}})& 42.24 $\rightarrow$ 68.12({\color{OliveGreen}{$\uparrow$25.88}})\\
    31 -- 35 & 6.53 $\rightarrow$ 9.76 ({\color{LimeGreen}{$\uparrow$3.23}})& 9.28 $\rightarrow$ 14.95 ({\color{ForestGreen}{$\uparrow$5.67}})& 33.47 $\rightarrow$ 50.00 ({\color{OliveGreen}{$\uparrow$16.53}})& 37.64 $\rightarrow$ 60.66({\color{OliveGreen}{$\uparrow$23.02}})\\
    35 -- 40 & 8.70 $\rightarrow$ 5.88 ({\color{WildStrawberry}{$\downarrow$2.82}})& 10.36 $\rightarrow$ 16.13 ({\color{ForestGreen}{$\uparrow$5.77}})& 29.57 $\rightarrow$ 20.00 ({\color{Mahogany}{$\downarrow$9.57}})& 34.83 $\rightarrow$ 54.22({\color{OliveGreen}{$\uparrow$19.39}})\\
    \bottomrule
    \end{tabular}}
    \caption{Female scholars' and male scholars' academic dropout rate (no paper published since 2018 Jan), given the total number of people of that gender and total number of people of that gender in the industry. Female scholars and male scholars industry rates given the total number of people of that gender and the total number of people of that gender who have dropped out.}
    \label{tab:domancy_table_with_change}
\end{table*}

\section{Additional Analysis of Clusters}

\subsection{Machine-Identified 9 Clusters}\label{appd:machine-identified}
\begin{figure*}[t]
\minipage{0.32\textwidth}
  \includegraphics[width=\linewidth]{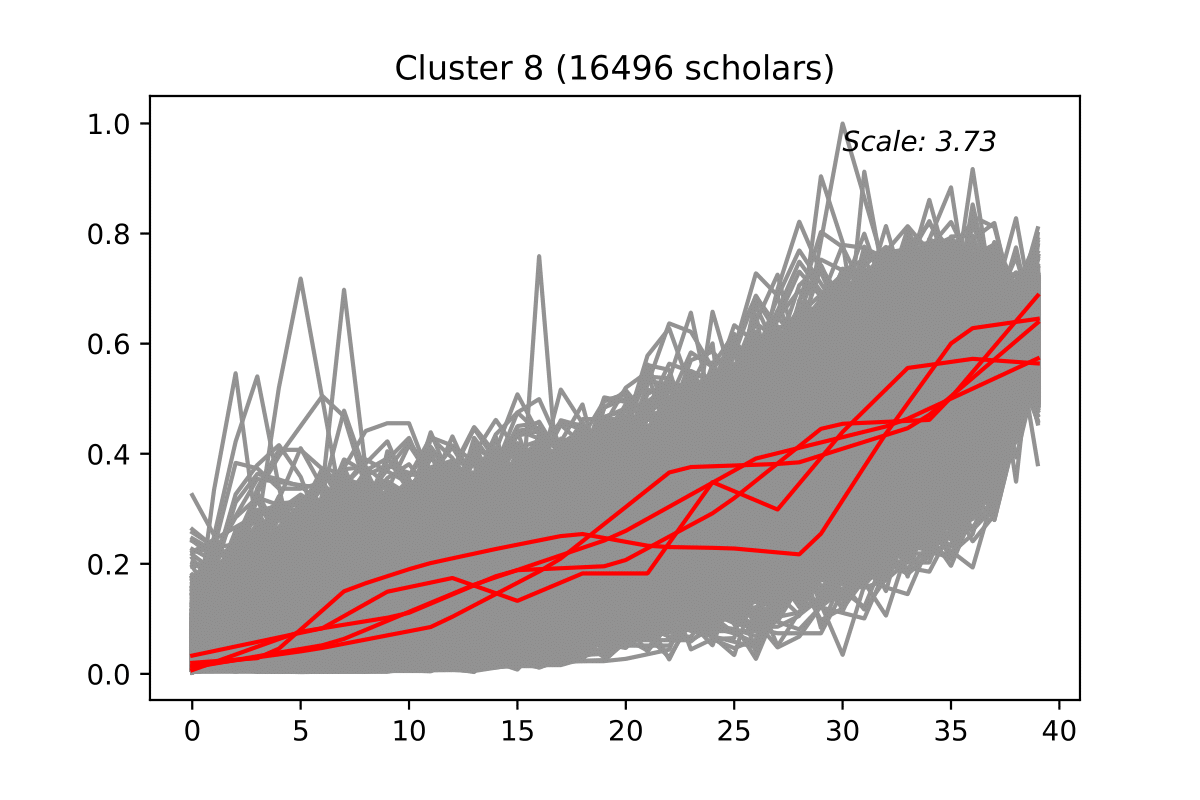}
\endminipage\hfill
\minipage{0.32\textwidth}
  \includegraphics[width=\linewidth]{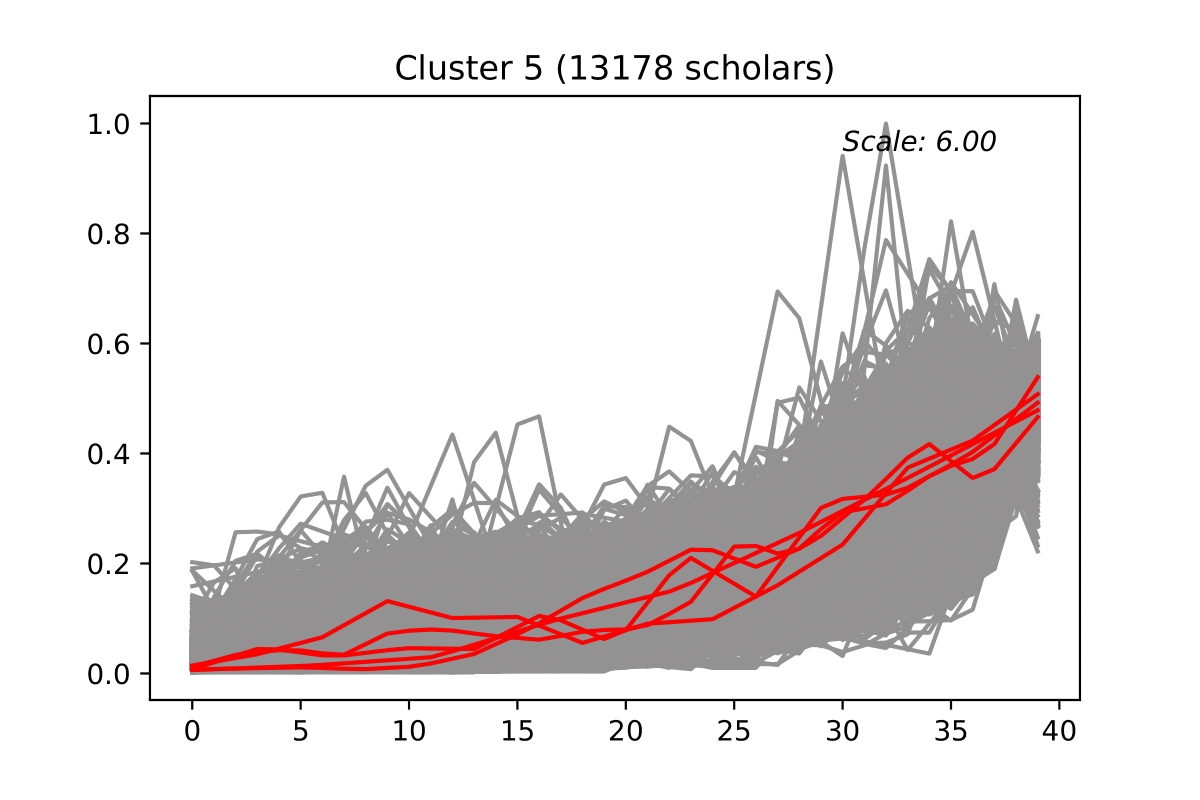}
\endminipage\hfill
\minipage{0.32\textwidth}%
  \includegraphics[width=\linewidth]{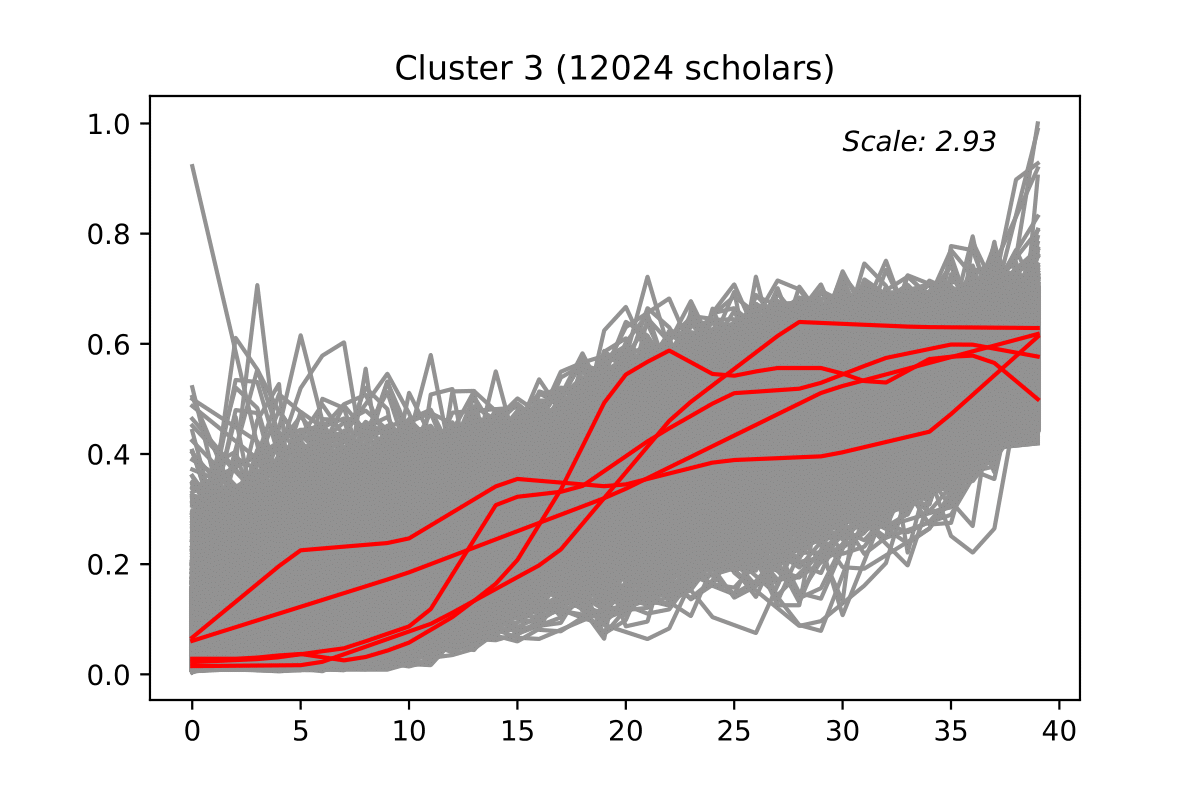}
\endminipage\hfill
\caption{Linear}\label{cluster_linear}
\minipage{0.32\textwidth}
  \includegraphics[width=\linewidth]{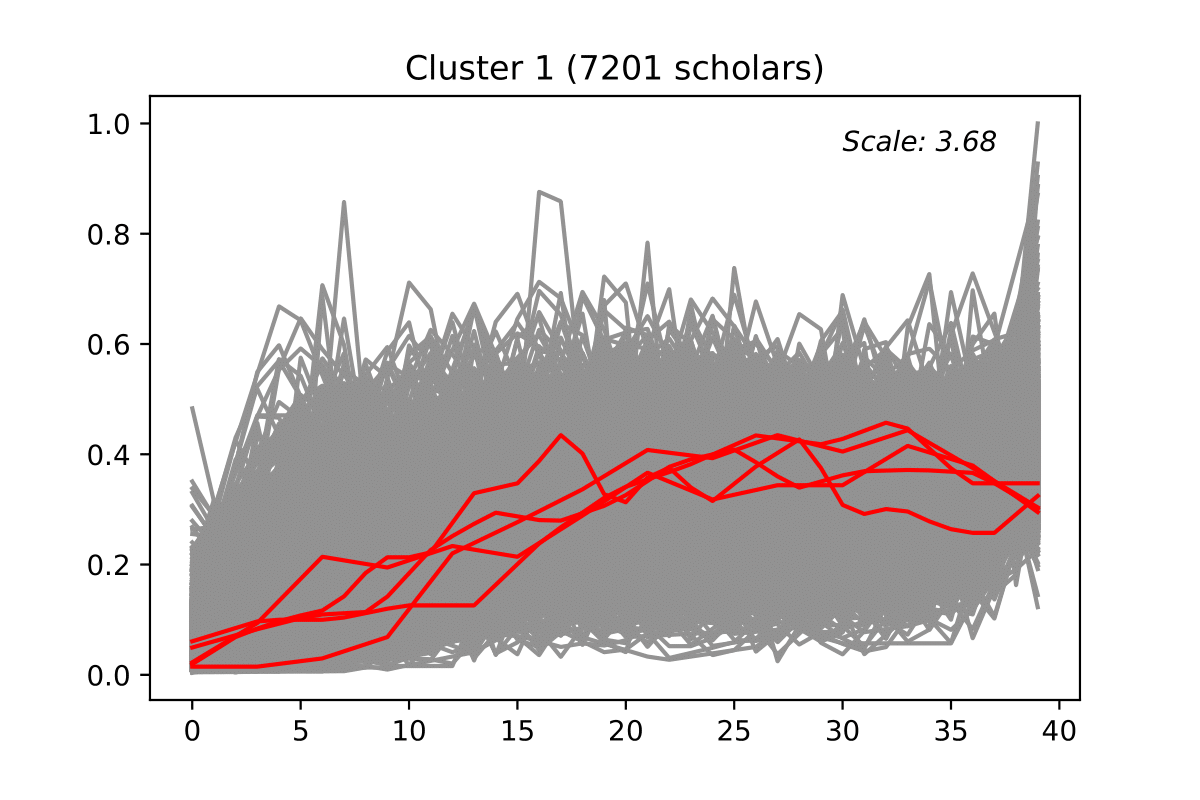}
\endminipage\hfill
\minipage{0.32\textwidth}
  \includegraphics[width=\linewidth]{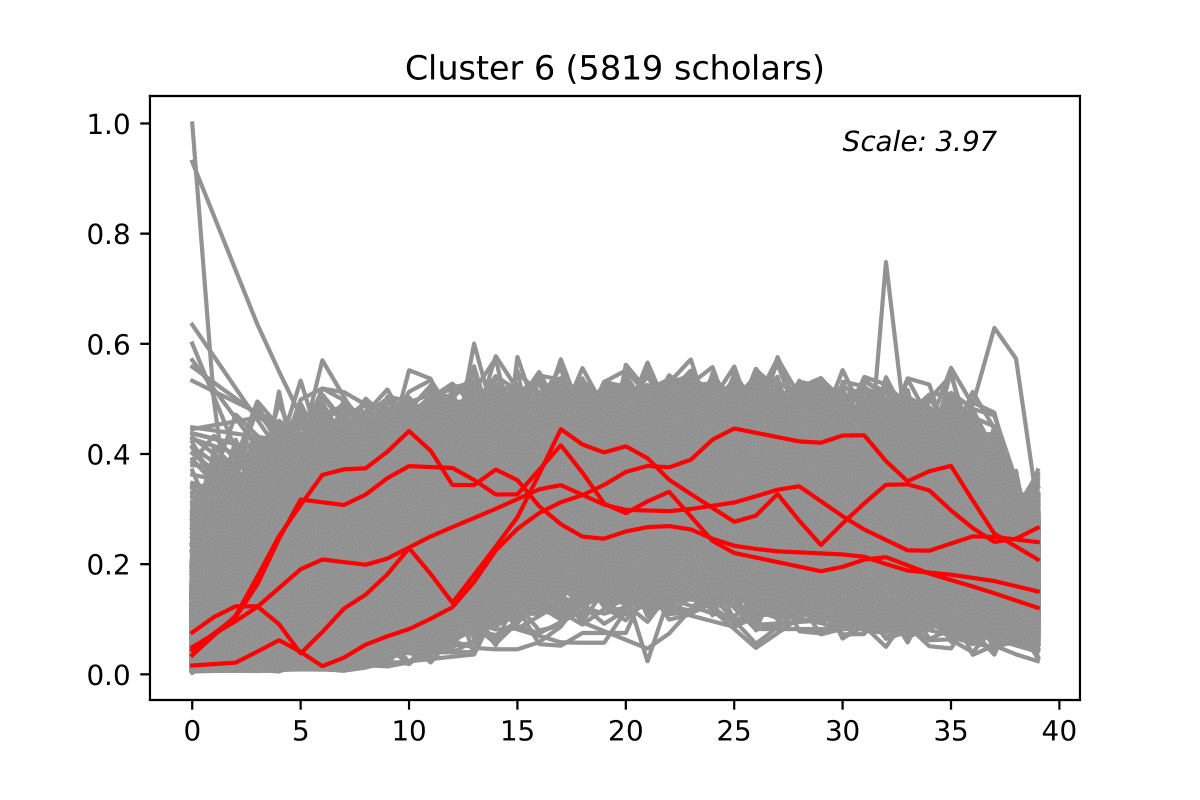}
\endminipage\hfill
\minipage{0.32\textwidth}%
  \includegraphics[width=\linewidth]{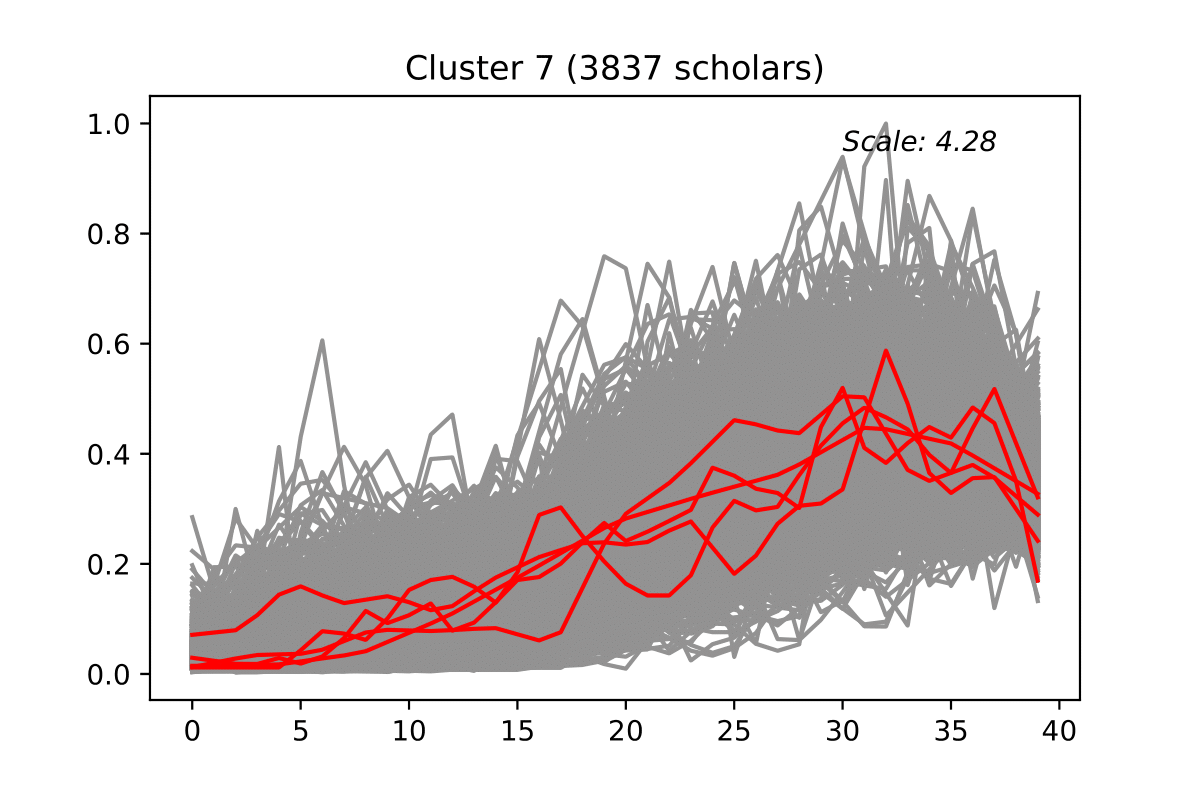}
\endminipage\hfill
\caption{Stumbling}\label{cluster_stumbling}
\minipage{0.32\textwidth}
  \includegraphics[width=\linewidth]{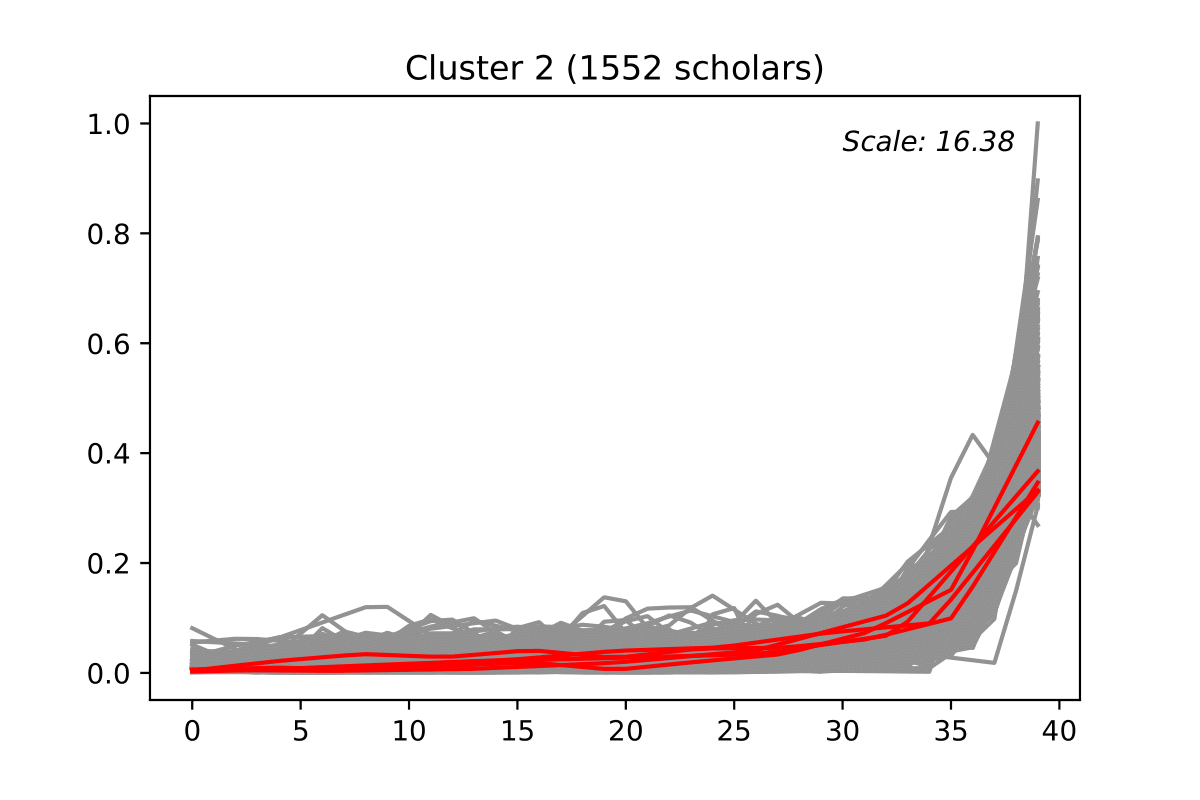}
\endminipage\hfill
\minipage{0.32\textwidth}
  \includegraphics[width=\linewidth]{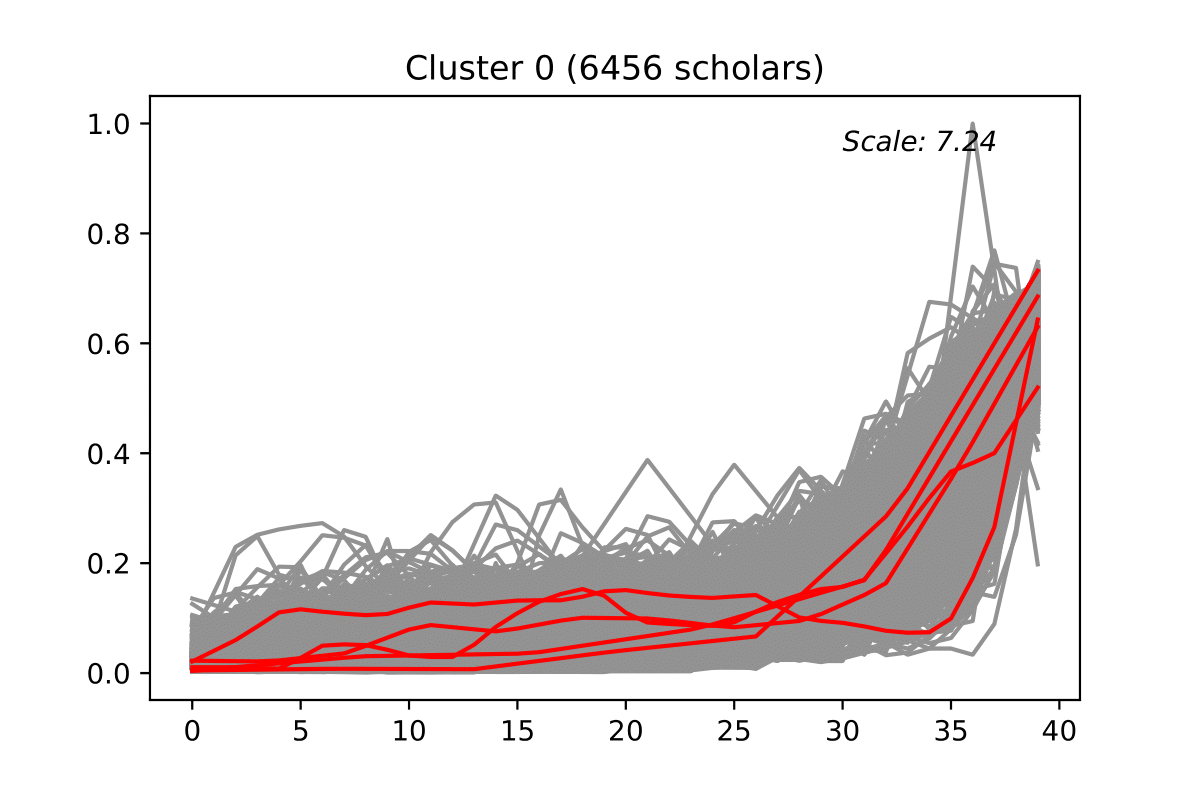}
\endminipage
\caption{Exponential}\label{cluster_exp}
\minipage{0.32\textwidth}%
  \includegraphics[width=\linewidth]{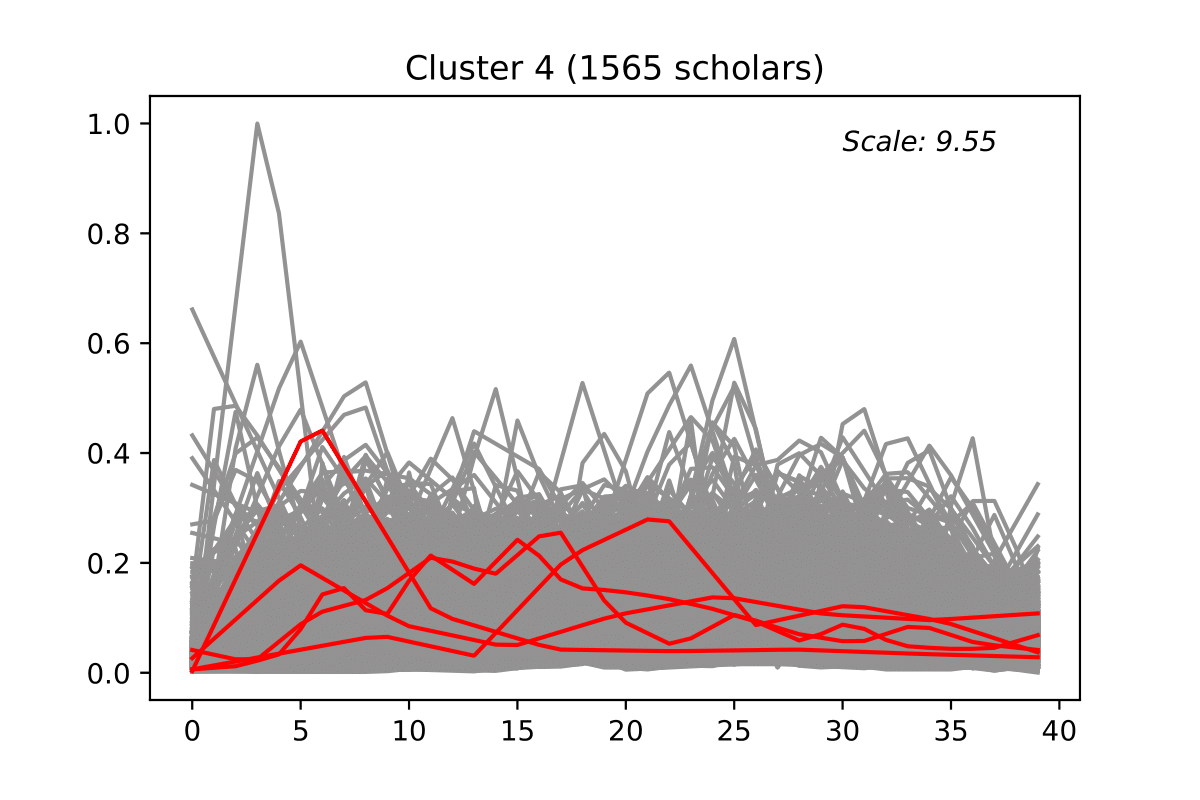}
\endminipage
\caption{Struggling}\label{cluster_struggling}
\caption{Citation trend of 78k scholars over their active years in 9 clusters. We further manually group them into 4 types. Clusters in \cref{cluster_linear} show linear growth, \cref{cluster_stumbling} shows a common trend from rise to decline, \cref{cluster_exp} shows exponential growth, and clusters in \cref{cluster_struggling} is struggling. Each grey line represents the trend of an AI scholar, and five of them are randomly sampled and labeled red for easy reading. The title of each plot contains the number of scholars in that cluster. The plot also labels the scale for each cluster.}
\label{fig:clusters_citation_by_year_appendix}
\end{figure*}

\cref{fig:clusters_citation_by_year_appendix} plots all 9 clusters generated by the TimeSeriesKMeans method. As we can see from these 9 clusters, cluster 3, 5, and 8 shows the linear growth pattern of citation; cluster 1, 6, and 7 shows a common trend from rising to decline; cluster 2 and 6 shows the exponential growth; and cluster 9 shows the struggling trend of citation. Thus we manually group them into 4 general patterns and select representative cluster in \cref{fig:clusters_citation_by_year}.





\subsection{NLP Scholars and Time Series Cluster}
The data of average academic ages per cluster is in \cref{tab:cluster_nlp}. With the cluster label and academic age for each scholar, the Pearson coefficient is -0.039 and the p-value is 3.72e-24 using Pearson's $\chi^2$ test. Therefore, it shows a strong correlation between the cluster a scholar belongs to and their academic age.

\begin{table}[H]
    \centering
    \small
    \resizebox{\columnwidth}{!}{
    \begin{tabular}{lcccc}
    \toprule
    & Exponential & Stumbling & Linear & Struggling  \\    \midrule
    F. NLP / F. Total & 137 / 718 & 678 / 3830 & 311 / 1534 & 32 / 131\\
    M. NLP / M. Total & 332 / 3028 & 2039 / 16565 & 1041 / 7640 & 79 / 728 \\
    F.\% ratio in NLP & 29.21 & 24.95 & 23.00 & 28.83 \\
    F. NLP academic age & 17.50$\pm$9.58 & 16.04$\pm$10.56 & 19.95$\pm$9.83 & 21.75$\pm$8.71  \\
    M. NLP academic age & 18.76$\pm$10.88 & 16.45$\pm$11.20 & 21.43$\pm$10.87 & 24.18$\pm$12.13 \\
    F. academic age & 16.98$\pm$9.12 & 14.74$\pm$9.80 & 18.40$\pm$9.24 & 19.92$\pm$8.18 \\ %
    M. academic age & 18.75$\pm$11.50 & 15.72$\pm$10.78 & 20.66$\pm$10.99 & 23.54$\pm$10.92 \\ %
    F. NLP citation & 1884$\pm$4771 & 2002$\pm$3867 & 2244$\pm$9023 & 392$\pm$375 \\
    M. NLP citation & 2413$\pm$4051 & 2841$\pm$7565 & 1996$\pm$4175 & 860$\pm$2242 \\
    F. citation & 1569$\pm$3826 & 1897$\pm$5545 & 1895$\pm$9233 & 438$\pm$698 \\ %
    M. citation & 3398$\pm$15660 & 2939$\pm$11550 & 2660$\pm$10068 & 827$\pm$2332 \\ %
    \bottomrule
    \end{tabular}
    }
    \caption{NLP scholars count and \# female\_nlp / (\# female\_nlp + \# male\_nlp) for each cluster. Exponential growth clusters have a larger Female ratio in NLP. Female NLP scholars also have higher average citations than nearly all female scholars in all fields, while male scholars are not.}
    \label{tab:cluster_nlp}
\end{table}


\section{Additional Analysis of Coauthorship}
We plot a heatmap of statistics in \cref{tab:paper_author_relation} for better visualization.

\begin{figure}[H]
    \centering
    \includegraphics[width=\columnwidth]{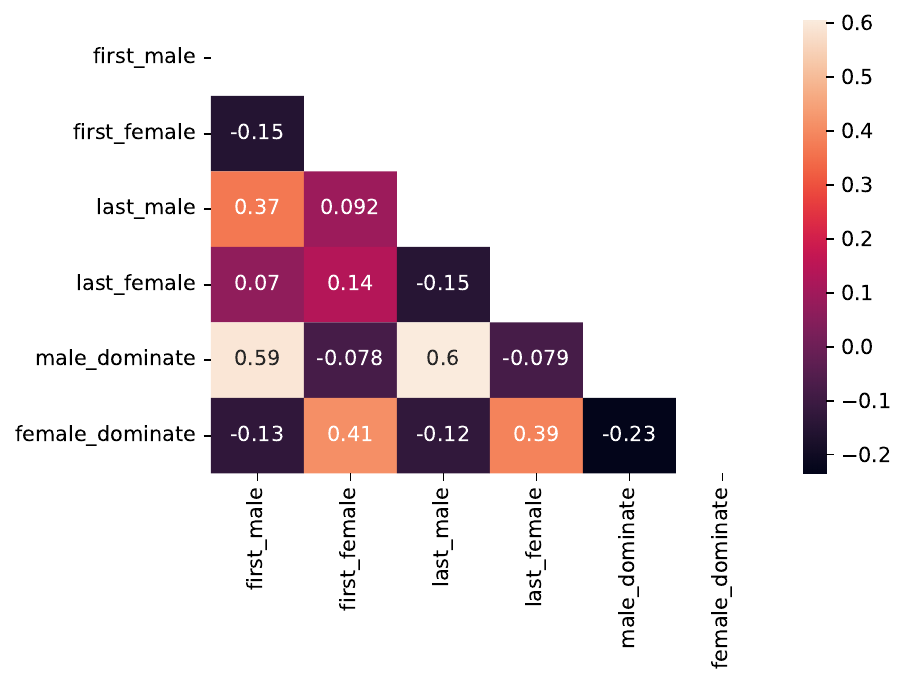}
    \caption{Heatmap corresponding to table \ref{tab:paper_author_relation}. The dataframe is constructed using boolean value. There are 91790 papers which includes unclassified gender paper.
}
    \label{fig:unclassified_heatmap}
\end{figure}

\section{Linguist Style}
\subsection{General linguistic statistics}\label{appd:linguistic}
We follow the same features set of \citet{jin2022ai} but extend their linguistic analysis in terms of stylish titles and separation of female- or male-first author papers.

\subsection{Stylish Titles} \label{appd:stylish_title}

The algorithm for detecting stylish paper titles is in \cref{alg:title_rules}. We use Part of Speech tags to capture certain syntactic characteristics. Despite its simplicity, the detection result fairly conforms with human labels of stylish titles in our understanding. \cref{tab:sorttopsty} and \cref{tab:sorttopsty_female} get the top 10 male scholars and top 5 female scholars that have the most number of stylish paper titles, and show their stylish title examples.




\begin{algorithm}[H]
    \caption{Algorithms of the stylish title detector}\label{alg:title_rules}
    \begin{algorithmic}
        \REQUIRE{The title of a paper}
        \ENSURE{A boolean whether the title is stylish}
        \STATE title remove noises and convert to lowercase
        \IF{special punctuation in title}
            \STATE return True
        \ELSIF{1st or 2nd personal pronouns in title}
            \STATE return True
        \ELSIF{meaningful numeric values in title}
            \STATE return True
        \ENDIF
        \STATE return False
    \end{algorithmic}
\end{algorithm}


\begin{table*}[t]
    \centering
    \small
    \begin{tabularx}{\textwidth}{cccX}
    \toprule
    Scholar name & \# Stylish titles & Portion (\%) & Sample titles \\
    \midrule
    \multirow{3}{*}{
    T. Y. W. (M)} & \multirow{3}{*}{294} & \multirow{3}{*}{28.25\%} & Does the Photographic Angle of Incidence Alter the Measured Fractal Dimension of the Retinal Vasculature? \\
    & & & We can save not only lives, but also quality of life: submandibular gland-sparing neck dissection \\
    & & & Erratum to: Is Sensory Loss an Understudied Risk Factor for Frailty? A Systematic Review and Meta-analysis \\
    \hline
    \multirow{3}{*}{
    F. M. (M)} & \multirow{3}{*}{243} & \multirow{3}{*}{22.62\%} & A systematic review of solid-pseudopapillary neoplasms: Are these rare lesions? \\
    & & & CT during arterial portography for the preoperative evaluation of hepatic tumors: how, when, and why? \\
    & & & Bikeshare: Barriers, facilitators and impacts on car use \\
    \hline
    \multirow{3}{*}{
    M. P. (M)} & \multirow{3}{*}{236} & \multirow{3}{*}{43.89\%} & Breaking the spell: Religion as a natural phenomenon \\
    & & & Are we explaining consciousness yet? \\
    & & & Speaking for our selves: An assessment of multiple personality disorder \\
    \hline
    \multirow{3}{*}{
    S. G. (M)} & \multirow{3}{*}{236} & \multirow{3}{*}{31.81\%} & Is baseline autonomic tone associated with new onset atrial fibrillation?: Insights from the framingham heart study \\
    & & & Biventricular pacing: more is better! \\
    & & & 6 Field evaluation of insecticides and neem formulations for management of brinjal shoot and fruit borer, Leucinodes orbonalis Guenee in brinjal \\
    \hline
    \multirow{3}{*}{
    J. B. (M)} & \multirow{3}{*}{234} & \multirow{3}{*}{29.83\%} & Energy, EROI and quality of life \\
    & & & Integrated child development services (ICDS) scheme: a journey of 37 years \\
    & & & Two methods for load balanced distributed adaptive integration \\
    \hline
    \multirow{3}{*}{
    D. D. R. (M)} & \multirow{3}{*}{222} & \multirow{3}{*}{28.41\%} & Information Power Grid: The new frontier in parallel computing? \\
    & & & Depth-first vs. best-first search \\
    & & & Top 10 algorithms in data mining. Survey paper \\
    \hline
    \multirow{3}{*}{
    J. C. (M)} & \multirow{3}{*}{219} & \multirow{3}{*}{27.63\%} & Do LGBT workplace diversity policies create value for firms? \\
    & & & Peering vs. transit: Performance comparison of peering and transit interconnections \\
    & & & 4 Strong Association Between the-308 TNF Promoter Polymorphism and Allergic Rhinitis in Pakistani Patients \\
    \hline
    \multirow{3}{*}{
    G. K. (M)} & \multirow{3}{*}{218} & \multirow{3}{*}{47.63\%} & Working knowledge: How organizations manage what they know \\
    & & & Thinking for a living: How to get better performances and results from knowledge workers \\
    & & & Saving IT's soul: Human-centered information management. \\
    \hline
    \multirow{3}{*}{
    J. H. (M)} & \multirow{3}{*}{212} & \multirow{3}{*}{29.67\%} & Pharmacotherapy-based problems in the management of diabetes mellitus: Needs much more to be done! \\
    & & & Long run relationship between gold prices, oil prices and Karachi stock market \\
    & & & Dengue fever again in Pakistan: Are we going in the right direction \\
    \hline
    \multirow{3}{*}{
    T. D. (M)} & \multirow{3}{*}{210} & \multirow{3}{*}{31.27\%} & Cloning, characterization and localization of a novel basic peroxidase gene from Catharanthus roseus \\
    & & & Technology Packages: Solar, biomass and hybrid dryers \\
    & & & Spinal tuberculosis with concomitant spondylolisthesis: coexisting entities or ‘cause and effect’? \\
    \bottomrule
    \end{tabularx}
    \caption{Top 10 male scholars sorted by the number of stylish titles.}
    \label{tab:sorttopsty}
\end{table*}

\begin{table*}[t]
    \centering
    \small
    \begin{tabularx}{\textwidth}{cccX}
    \toprule
    Scholar name & \# Stylish titles & Portion (\%) & Sample titles \\
    \midrule
    \multirow{3}{*}{
    P. V. (F)} & \multirow{3}{*}{155} & \multirow{3}{*}{44.87\%} & ‘Spam, spam, spam, spam... Lovely spam!’ Why is Bluespam different? \\
    & & & One world one dream? Sports blogging at the Beijing Olympic Games \\
    & & & Forget me (in Europe), forget me not (outside Europe) \\
    \hline
    \multirow{3}{*}{
    A. P. (F)} & \multirow{3}{*}{130} & \multirow{3}{*}{27.86\%} & Is the wolf angry or... just hungry? \\
    & & & Tell me that bit again... bringing interactivity to a virtual storyteller \\
    & & & ``I want to slay that Dragon!'' -- Influencing choice in interactive storytelling \\
    \hline
    \multirow{3}{*}{
    K. D. (F)} & \multirow{3}{*}{123} & \multirow{3}{*}{33.55\%} & How may I serve you?: A robot companion approaching a seated person in a helping context \\
    & & & I could be you: The phenomenological dimension of social understanding \\
    & & & Robots we like to live with! A developmental perspective on a personalized, life-long robot companion \\
    \hline
    \multirow{3}{*}{
    M. H. (F)} & \multirow{3}{*}{99} & \multirow{3}{*}{46.73\%} & Defining profiling: A new type of knowledge? \\
    & & & Location Data, Purpose Binding and Contextual Integrity: What’s the Message? \\
    & & & Dualism is dead. Long live plurality (instead of duality) \\
    \hline
    \multirow{3}{*}{
    P. S. (F)} & \multirow{3}{*}{95} & \multirow{3}{*}{24.81\%} & Does Your Food Affect Your Intelligence? \\
    & & & Taking leads out of nature, can nano deliver us from COVID-like pandemics? \\
    & & & Biting off more than we could chew -- A surprising find on biopsy! \\
    \bottomrule
    \end{tabularx}
    \caption{Top 5 female scholars sorted by the number of stylish titles.}
    \label{tab:sorttopsty_female}
\end{table*}

\subsection{Full List of LIWC Features}\label{appd:liwc}

\cref{tab:liwc1} and \cref{tab:liwc2} show a full list of word categories along with their frequency by using LIWC 2015 \cite{pennebaker2001linguistic}.

\begin{table*}[t]
    \centering
    \small
    \resizebox{\textwidth}{!}{\begin{tabular}{llccccc}
    \toprule
    Category & & \textbf{Score (All)} & \multicolumn{3}{c}{\textbf{Score (Female Abstracts)}} \\
    \cline{4-6}
    & & & 1st=F & $\geq$50\% F & Last=F \\ \midrule
    Word Count \\ \midrule
    \multicolumn{3}{l}{\textbf{\textit{Summary Language Variables}}} \\
Words/Sentence &  &  &  &  \\
Words > 6 Letters &  &  &  &  \\
\multicolumn{3}{l}{	\textbf{\textit{Linguistic Dimensions}}} &  &  &  &  \\
{Total Function Words} & \{the, of, and, a, to\} & 43.91 $\pm$ 16.39  &  45.04 $\pm$ 16.30  &  44.43 $\pm$ 16.12  &  44.79 $\pm$ 16.73  \\
\hspace{1.5mm}{Total Pronouns} & \{that, this, we, which, it\} & 4.21 $\pm$ 3.06  &  4.39 $\pm$ 3.08  &  4.28 $\pm$ 3.05  &  4.48 $\pm$ 3.17  \\
\hspace{1.5mm}{Personal Pronouns} & \{we, our, they, them, us\} & 1.06 $\pm$ 1.38  &  1.17 $\pm$ 1.47  &  1.11 $\pm$ 1.41  &  1.19 $\pm$ 1.51  \\
\hspace{3mm}{1st Person Singular} & \{i, mine, my, im, me\} & 0.01 $\pm$ 0.16  &  0.01 $\pm$ 0.21  &  0.01 $\pm$ 0.19  &  0.01 $\pm$ 0.23  \\
\hspace{3mm}{1st Person Plural} & \{we, our, us, lets, ourselves\} & 0.90 $\pm$ 1.25  &  0.97 $\pm$ 1.28  &  0.93 $\pm$ 1.26  &  0.98 $\pm$ 1.31  \\
\hspace{3mm}{2nd Person} & \{you, your, u, ya, ye\} & 0.01 $\pm$ 0.17  &  0.01 $\pm$ 0.30  &  0.01 $\pm$ 0.25  &  0.01 $\pm$ 0.32  \\
\hspace{3mm}{3rd Person Singular} & \{his, her, he, she, him\} & 0.01 $\pm$ 0.17  &  0.01 $\pm$ 0.17  &  0.01 $\pm$ 0.15  &  0.01 $\pm$ 0.15  \\
\hspace{3mm}{3rd Person Plural} & \{they, them, themselves, their, theirs\} & 0.14 $\pm$ 0.42  &  0.16 $\pm$ 0.46  &  0.15 $\pm$ 0.45  &  0.17 $\pm$ 0.48  \\
\hspace{1.5mm}{Impersonal Pronouns} & \{that, this, which, it, these\} & 3.15 $\pm$ 2.31  &  3.22 $\pm$ 2.30  &  3.17 $\pm$ 2.28  &  3.29 $\pm$ 2.36  \\
\hspace{1.5mm}{Articles} & \{the, a an\} &10.27 $\pm$ 5.21  &  10.18 $\pm$ 5.21  &  10.00 $\pm$ 5.16  &  10.09 $\pm$ 5.20  \\
\hspace{1.5mm}{Prepositions} & \{of, to, in, for, with\} & 17.34 $\pm$ 6.86  &  18.02 $\pm$ 6.91  &  17.76 $\pm$ 6.87  &  17.81 $\pm$ 7.01  \\
\hspace{1.5mm}Auxiliary Verbs & \{is, are, be can, have\} & 5.58 $\pm$ 3.35  &  5.59 $\pm$ 3.31  &  5.62 $\pm$ 3.32  &  5.57 $\pm$ 3.40  \\
\hspace{1.5mm}{Common Adverbs} & \{such, also, when, only, where\} &1.88 $\pm$ 1.81  &  1.94 $\pm$ 1.82  &  1.86 $\pm$ 1.80  &  1.97 $\pm$ 1.87  \\
\hspace{1.5mm}{Conjunctions} & \{and, as, or, also, but\} & 5.64 $\pm$ 3.19  &  5.99 $\pm$ 3.26  &  5.92 $\pm$ 3.24  &  5.92 $\pm$ 3.26  \\
\hspace{1.5mm}{Negations} & \{not, without, no, cannot, negative\} & 0.36 $\pm$ 0.68  &  0.38 $\pm$ 0.71  &  0.38 $\pm$ 0.69  &  0.39 $\pm$ 0.70  \\
\multicolumn{3}{l}{	\textbf{\textit{Other Grammar}}} &  &  &  &  \\
\hspace{1.5mm}{Common Verbs} & \{is, are, be, using, based\} & 8.85 $\pm$ 4.65  &  9.01 $\pm$ 4.61  &  8.98 $\pm$ 4.63  &  8.99 $\pm$ 4.70  \\
\hspace{1.5mm}{Common Adjectives} &  \{as, different, new, more, than\} & 4.67 $\pm$ 3.13  &  4.87 $\pm$ 3.19  &  4.83 $\pm$ 3.17  &  4.88 $\pm$ 3.21  \\
\hspace{1.5mm}{Comparisons} & \{as, different, more, than, most\} & 2.43 $\pm$ 2.15  &  2.56 $\pm$ 2.21  &  2.55 $\pm$ 2.24  &  2.56 $\pm$ 2.22  \\
\hspace{1.5mm}{Interrogatives} & \{which, when, where, how, whether\} & 0.88 $\pm$ 1.08  &  0.91 $\pm$ 1.12  &  0.90 $\pm$ 1.11  &  0.95 $\pm$ 1.15  \\
\hspace{1.5mm}Numbers & \{two, one, first, three, single\} & 0.75 $\pm$ 1.20  &  0.76 $\pm$ 1.19  &  0.75 $\pm$ 1.14  &  0.74 $\pm$ 1.12  \\
\hspace{1.5mm}Quantifiers & \{more, each, both, most, all\} & 1.94 $\pm$ 1.87  &  1.95 $\pm$ 1.86  &  1.93 $\pm$ 1.84  &  1.93 $\pm$ 1.83  \\
\multicolumn{3}{l}{	\textbf{\textit{Psychological Processes}}} &  &  &  &  \\
{Affective Processes} & \{well, important, problems, energy, problem\} & 2.89 $\pm$ 2.46  &  2.98 $\pm$ 2.47  &  3.00 $\pm$ 2.53  &  3.02 $\pm$ 2.56  \\
\hspace{1.5mm}{Positive Emotion} & \{well, important, energy, better, support\} & 1.98 $\pm$ 1.88  &  2.05 $\pm$ 1.93  &  2.06 $\pm$ 1.96  &  2.11 $\pm$ 1.99  \\
\hspace{1.5mm}Negative Emotion & \{problems, problem, low, critical, difficult\} & 0.88 $\pm$ 1.39  &  0.88 $\pm$ 1.36  &  0.89 $\pm$ 1.40  &  0.86 $\pm$ 1.37  \\
\hspace{3mm}Anxiety & \{uncertainty, pressure, uncertainties, risk, risks\} &0.15 $\pm$ 0.57  &  0.15 $\pm$ 0.55  &  0.16 $\pm$ 0.58  &  0.14 $\pm$ 0.54  \\
\hspace{3mm}{Anger} & \{critical, attacks, argue, dominant, arguments\} &0.13 $\pm$ 0.54  &  0.15 $\pm$ 0.57  &  0.14 $\pm$ 0.53  &  0.15 $\pm$ 0.58  \\
\hspace{3mm}Sadness & \{low, lower, failure, missing, suffer\} & 0.20 $\pm$ 0.59  &  0.21 $\pm$ 0.62  &  0.21 $\pm$ 0.60  &  0.18 $\pm$ 0.54  \\
{Social Processes} & \{we, our, provide, they, provides\} & 2.81 $\pm$ 2.75  &  3.25 $\pm$ 3.06  &  3.13 $\pm$ 2.99  &  3.30 $\pm$ 3.04  \\
\hspace{1.5mm}{Family} & \{family, families, parents, pregnancy, son\} & 0.03 $\pm$ 0.25  &  0.03 $\pm$ 0.23  &  0.04 $\pm$ 0.32  &  0.03 $\pm$ 0.33  \\
\hspace{1.5mm}{Friends} & \{contact, neighborhood, neighboring, neighbors, date\} & 0.04 $\pm$ 0.29  &  0.04 $\pm$ 0.27  &  0.05 $\pm$ 0.31  &  0.05 $\pm$ 0.35  \\
\hspace{1.5mm}{Female References} & \{female, her, women, females, she\} & 0.01 $\pm$ 0.17  &  0.02 $\pm$ 0.21  &  0.02 $\pm$ 0.23  &  0.02 $\pm$ 0.21  \\
\hspace{1.5mm}{Male References} & \{his, male, he, men, son\} & 0.02 $\pm$ 0.22  &  0.02 $\pm$ 0.23  &  0.03 $\pm$ 0.24  &  0.02 $\pm$ 0.24  \\
    \bottomrule
    \end{tabular}
    }
    \caption{Linguistic features extracted by LIWC. Each number means occurrence per string (which is abstract).
    We also show the top 5 words from score-All.
    We compare features of general abstracts (using the 83K random sample), and features of abstracts of female-authored papers. Among female-authored papers, we analyze papers whose first author is female (1st=F), the last author is female (last=F), and over 50\% female authors.
    }
    \label{tab:liwc1}
\end{table*}

\begin{table*}[t]
    \centering
    \small
    \resizebox{\textwidth}{!}{
    \begin{tabular}{llccccc}
    \toprule
    Category &  & \textbf{Score (All)} & \multicolumn{3}{c}{\textbf{Score (Female Abstracts)}} \\
    \cline{4-6}
    & & & 1st=F & $\geq$50\% F & Last=F \\ \midrule
{{Cognitive Processes}} & \{using, based, or, used, results\} & 10.94 $\pm$ 5.81  &  11.37 $\pm$ 5.85  &  11.26 $\pm$ 5.85  &  11.59 $\pm$ 6.01  \\
\hspace{1.5mm}{{Insight}} & \{information, learning, analysis, knowledge, recognition\}  & 3.77 $\pm$ 2.95  &  4.06 $\pm$ 3.05  &  4.04 $\pm$ 3.04  &  4.17 $\pm$ 3.14  \\
\hspace{1.5mm}{Causation} & \{using, based, used, results, use\} & 3.44 $\pm$ 2.43  &  3.48 $\pm$ 2.44  &  3.45 $\pm$ 2.44  &  3.50 $\pm$ 2.45  \\
\hspace{1.5mm}{Discrepancy} & \{problems, problem, need, could, if\} & 0.56 $\pm$ 0.96  &  0.57 $\pm$ 0.94  &  0.57 $\pm$ 0.95  &  0.60 $\pm$ 0.99  \\
\hspace{1.5mm}Tentative & \{or, most, may, some, any\} & 1.72 $\pm$ 1.89  &  1.70 $\pm$ 1.84  &  1.68 $\pm$ 1.82  &  1.75 $\pm$ 1.87  \\
\hspace{1.5mm}{Certainty} & \{all, accuracy, specific, accurate, total\} & 0.88 $\pm$ 1.14  &  0.91 $\pm$ 1.14  &  0.88 $\pm$ 1.13  &  0.94 $\pm$ 1.21  \\
\hspace{1.5mm}{Differentiation} & \{or, different, not, than, other\} & 1.57 $\pm$ 1.69  &  1.65 $\pm$ 1.74  &  1.63 $\pm$ 1.73  &  1.69 $\pm$ 1.77  \\
Perceptual Processes & \{show, images, search, fuzzy, image\} & 1.42 $\pm$ 1.86  &  1.47 $\pm$ 1.88  &  1.42 $\pm$ 1.87  &  1.40 $\pm$ 1.86  \\
\hspace{1.5mm}See & \{show, images, search, image, shows\} & 0.85 $\pm$ 1.35  &  0.86 $\pm$ 1.32  &  0.85 $\pm$ 1.35  &  0.82 $\pm$ 1.30  \\
\hspace{1.5mm}{Hear} & \{noise, noisy, music, voice, speech\} & 0.17 $\pm$ 0.74  &  0.21 $\pm$ 0.81  &  0.19 $\pm$ 0.75  &  0.20 $\pm$ 0.80  \\
\hspace{1.5mm}{Feel} & \{fuzzy, flexible, weight, weighted, hand\} & 0.25 $\pm$ 0.80  &  0.23 $\pm$ 0.78  &  0.23 $\pm$ 0.75  &  0.23 $\pm$ 0.77  \\
{Biological Processes} & \{clinical, expression, face, medical, physical\} &1.16 $\pm$ 2.20  &  1.37 $\pm$ 2.41  &  1.44 $\pm$ 2.54  &  1.18 $\pm$ 2.30  \\
\hspace{1.5mm}Body & \{face, blood, hand, heart, neurons\} & 0.28 $\pm$ 0.98  &  0.31 $\pm$ 1.04  &  0.31 $\pm$ 1.03  &  0.25 $\pm$ 0.93  \\
\hspace{1.5mm}{Health} & \{clinical, medical, physical, health, diagnosis\} & 0.71 $\pm$ 1.67  &  0.85 $\pm$ 1.83  &  0.92 $\pm$ 1.95  &  0.76 $\pm$ 1.80  \\
\hspace{1.5mm}{Sexual} & \{prostate, pregnancy, sex, ovarian, arousal\} & 0.02 $\pm$ 0.25  &  0.02 $\pm$ 0.32  &  0.02 $\pm$ 0.28  &  0.03 $\pm$ 0.33 \\
\hspace{1.5mm}{Ingestion} & \{expression, water, weight, expressions, expressed\} & 0.16 $\pm$ 0.68  &  0.19 $\pm$ 0.75  &  0.20 $\pm$ 0.81  &  0.16 $\pm$ 0.67  \\
{Drives} & \{we, approach, our, first, over\} & 6.65 $\pm$ 4.26  &  6.92 $\pm$ 4.21  &  6.82 $\pm$ 4.27  &  7.07 $\pm$ 4.40  \\
\hspace{1.5mm}{Affiliation} & \{we, our, social, communication, interaction\} & 1.62 $\pm$ 1.92  &  1.81 $\pm$ 2.03  &  1.76 $\pm$ 2.00  &  1.84 $\pm$ 2.03  \\
\hspace{1.5mm}{Achievement} &  \{first, work, efficient, obtained, better\} & 2.15 $\pm$ 1.99  &  2.19 $\pm$ 1.97  &  2.15 $\pm$ 1.98  &  2.23 $\pm$ 2.05  \\
\hspace{1.5mm}{Power} & \{over, high, order, large, important\} & 2.11 $\pm$ 2.06  &  2.15 $\pm$ 2.07  &  2.15 $\pm$ 2.08  &  2.21 $\pm$ 2.12  \\
\hspace{1.5mm}{Reward} & \{approach, obtained, approaches, better, best\} & 1.10 $\pm$ 1.30  &  1.11 $\pm$ 1.29  &  1.11 $\pm$ 1.30  &  1.16 $\pm$ 1.36  \\
\hspace{1.5mm}Risk & \{problems, problem, security, difficult, lack\} & 0.52 $\pm$ 1.02  &  0.50 $\pm$ 0.97  &  0.51 $\pm$ 1.01  &  0.50 $\pm$ 1.00  \\
Time Orientations &  &  &  &  \\
\hspace{1.5mm}{Past Focus} & \{used, was, been, were, obtained\} & 1.96 $\pm$ 2.23  &  2.14 $\pm$ 2.39  &  2.19 $\pm$ 2.45  &  2.01 $\pm$ 2.25  \\
\hspace{1.5mm}Present Focus & \{is, are, be, can, have\} & 6.29 $\pm$ 3.65  &  6.26 $\pm$ 3.63  &  6.16 $\pm$ 3.65  &  6.37 $\pm$ 3.71  \\
\hspace{1.5mm}Future Focus & \{may, then, will, prediction, future\} & 0.61 $\pm$ 1.08  &  0.60 $\pm$ 1.05  &  0.63 $\pm$ 1.09  &  0.65 $\pm$ 1.11  \\
{Relativity} & \{in, on, at, approach, new\} & 10.84 $\pm$ 5.64  &  11.09 $\pm$ 5.66  &  11.00 $\pm$ 5.61  &  10.95 $\pm$ 5.70  \\
\hspace{1.5mm}Motion & \{approach, approaches, behavior, changes, increase\} & 1.44 $\pm$ 1.62  &  1.47 $\pm$ 1.61  &  1.44 $\pm$ 1.61  &  1.47 $\pm$ 1.63  \\
\hspace{1.5mm}{Space} & \{in, on, at, into, both\} & 6.96 $\pm$ 4.05  &  7.10 $\pm$ 4.07  &  7.05 $\pm$ 4.03  &  6.98 $\pm$ 4.03  \\
\hspace{1.5mm}Time & \{new, present, first, when, then\} & 2.40 $\pm$ 2.20  &  2.44 $\pm$ 2.25  &  2.44 $\pm$ 2.23  &  2.43 $\pm$ 2.23  \\
Personal Concerns &  &  &  &  \\
\hspace{1.5mm}{Work} & \{performance, learning, analysis, paper, applications\} & 4.53 $\pm$ 3.58  &  4.78 $\pm$ 3.70  &  4.74 $\pm$ 3.69  &  4.95 $\pm$ 3.91  \\
\hspace{1.5mm}Leisure & \{novel, expression, channels, videos, play\} & 0.48 $\pm$ 1.01  &  0.51 $\pm$ 1.08  &  0.51 $\pm$ 1.06  &  0.48 $\pm$ 1.00  \\
\hspace{1.5mm}Home & \{address, family, home, neighborhood, neighboring\} & 0.12 $\pm$ 0.45  &  0.11 $\pm$ 0.44  &  0.12 $\pm$ 0.44  &  0.12 $\pm$ 0.47  \\
\hspace{1.5mm}Money & \{investigate, cost, investigated, free, economic\} & 0.42 $\pm$ 1.02  &  0.43 $\pm$ 0.99  &  0.42 $\pm$ 1.00  &  0.44 $\pm$ 1.03  \\
\hspace{1.5mm}{Religion} & \{beliefs, moral, sacrificing, monkeys, agnostic\} & 0.01 $\pm$ 0.17  &  0.02 $\pm$ 0.18  &  0.02 $\pm$ 0.19  &  0.02 $\pm$ 0.21  \\
\hspace{1.5mm}{Death} & \{mortality, die, mortality, deaths, death\} & 0.04 $\pm$ 0.31  &  0.04 $\pm$ 0.32  &  0.04 $\pm$ 0.36  &  0.04 $\pm$ 0.33  \\
Informal Language & \{well, o, da, en, um\} & 0.16 $\pm$ 0.66  &  0.16 $\pm$ 0.65  &  0.16 $\pm$ 0.68  &  0.15 $\pm$ 0.64  \\
\hspace{1.5mm}Swear Words & \{retardation, dummy, screws, screw, retarded\} & 0.00 $\pm$ 0.05  &  0.00 $\pm$ 0.04  &  0.00 $\pm$ 0.03  &  0.00 $\pm$ 0.03  \\
\hspace{1.5mm}{Netspeak} & \{o, da, em, k, mm\} & 0.04 $\pm$ 0.50  &  0.04 $\pm$ 0.53  &  0.04 $\pm$ 0.53  &  0.04 $\pm$ 0.51  \\
\hspace{1.5mm}Assent & \{k, indeed, agree, absolutely, cool\} &0.01 $\pm$ 0.13  &  0.01 $\pm$ 0.15  &  0.01 $\pm$ 0.13  &  0.01 $\pm$ 0.12  \\
\hspace{1.5mm}Nonfluencies & \{well, um, mm, er, ah\} & 0.11 $\pm$ 0.37  &  0.12 $\pm$ 0.37  &  0.11 $\pm$ 0.36  &  0.11 $\pm$ 0.35  \\
\hspace{1.5mm}Fillers & \{rrani, rranr\} & 0.00 $\pm$ 0.01  &  0.00 $\pm$ 0.00  &  0.00 $\pm$ 0.00  &  0.00 $\pm$ 0.00  \\
    \bottomrule
    \end{tabular}
    }
    \caption{Following \cref{tab:liwc1}.
    \label{tab:liwc2}
    }
\end{table*}

\end{document}